\documentclass[runningheads]{llncs}

\usepackage[T1]{fontenc}

\usepackage{cite}
\usepackage{comment}
\usepackage{amsmath}
\usepackage{amssymb}
\usepackage{mathtools}
\usepackage{amsfonts}       
\usepackage{subcaption}
\usepackage{booktabs} 
\usepackage{hyperref}
\usepackage{nicefrac}       
\usepackage{microtype}      
\usepackage{xcolor}         
\usepackage{graphicx}
\usepackage[labelfont=bf]{caption}
\usepackage{float}
\usepackage{booktabs}
\usepackage{multirow}
\usepackage{acro}
\usepackage[export]{adjustbox}[2011/08/13]
\usepackage{tikz}
\usetikzlibrary{decorations.pathreplacing}
\usepackage{placeins}

\DeclareAcronym{ours}{
  short=CLL,
  long=Calibrated Lipschitz-Margin Loss,
}
\newcommand{\spow}[1]{{\scalebox{0.5}{$(#1)$}}}
\newcommand{\kf}{K^{\scalebox{0.5}{$(f)$}}}
\newcommand{\khf}{K^{\scalebox{0.5}{$(h \circ f)$}}}
\newcommand{\kh}{K^{\scalebox{0.5}{$(h)$}}}
\newcommand{\maxkf}{\overline{K}^\spow{f}}

\newcommand{\maxkhf}{\overline{K}^\spow{h\circ f}}

\newcommand{\myparagraph}[1]{\textbf{#1}}

\def\eg{{e.g.}} \def\Eg{{E.g.}}
\def\ie{{i.e.}} \def\Ie{{I.e.}}

\def\wrt{{w.r.t.~}}

\definecolor{legendred}{rgb}{0.72,0.26,0.27}
\definecolor{legendgreen}{rgb}{0.29,0.6,0.38}
\definecolor{legendblue}{rgb}{0.24,0.41,0.64}
\definecolor{legendbck}{rgb}{0.47,0.47,0.47}

\newif\ifreview
\reviewfalse

\ifreview
    \usepackage{lineno}

    \linenumbers
\fi

\begin{document}


\def\SubNumber{087}

\def\GCPRTrack{Fast Review Track}

\title{Certified Robust Models with Slack Control and Large Lipschitz Constants}

\ifreview
\else

    \author{Max Losch\inst{1} \and
    David Stutz\inst{1} \and
    Bernt Schiele\inst{1} \and
    Mario Fritz\inst{2} 
    }
    
    \authorrunning{M. Losch et al.}
    
    \institute{$^1$Max Planck Institute for Informatics, Saarland Informatics Campus\\ Saarbr\"ucken, Germany\\
    \email{\{mlosch, dstutz, schiele\}@mpi-inf.com}\\
    $^2$CISPA Helmholtz Center for Information Security\\
    Saarbr\"ucken, Germany\\
    \email{fritz@cispa.de}}
\fi

\maketitle              

\begin{abstract}
Despite recent success, state-of-the-art learning-based models remain highly vulnerable to input changes such as adversarial examples.
In order to obtain certifiable robustness against such perturbations, recent work considers Lipschitz-based regularizers or constraints while at the same time increasing prediction margin. 
Unfortunately, this comes at the cost of significantly decreased accuracy.
In this paper, we propose a \ac{ours}
that addresses this issue and improves certified robustness by tackling two problems:
Firstly, commonly used margin losses do not adjust the penalties to the shrinking output distribution; caused by minimizing the Lipschitz constant $K$.
Secondly, and most importantly, we observe that  minimization of $K$ can lead to overly smooth decision functions.
This limits the model's complexity and thus reduces accuracy. 
Our \ac{ours} addresses these issues by explicitly calibrating the loss \wrt{margin} and Lipschitz constant, thereby establishing full control over slack and improving robustness certificates even with larger Lipschitz constants.
On CIFAR-10, CIFAR-100 and Tiny-ImageNet, our models consistently outperform losses that leave the constant unattended.
On CIFAR-100 and Tiny-ImageNet, \ac{ours} improves upon state-of-the-art deterministic $L_2$ robust accuracies.
In contrast to current trends, we unlock potential of much smaller models without $K$=$1$ constraints.

\keywords{Certified robustness, Lipschitz, Large margin, Slack}
\end{abstract}
\section{Introduction}

\begin{figure}
    \centering
    \includegraphics[width=0.5\linewidth]{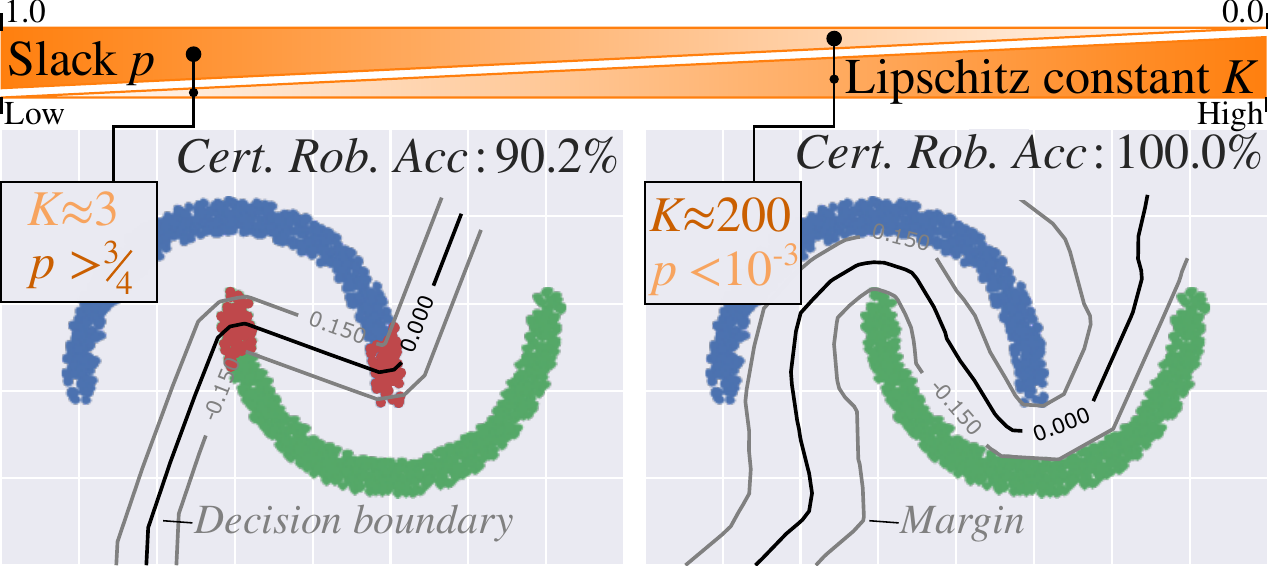}
    \vspace*{-1em}
    \caption{Existing Lipschitz margin methods control the Lipschitz constant $K$ to be low, yet we observe decision functions becoming overly smooth when $K$ is too low (left) -- impairing accuracy. Our \ac{ours} loss provides slack control, which we show is inversely proportional to $K$ (see gradients on top). We can control $K$ to be high, avoid the smoothing and achieve improved clean and robust accuracies (right). Incorrect or not robust samples \textcolor{legendred}{marked red}.
    }
    \label{fig:margin}
\end{figure}
The Lipschitz constant $K$ of a classifier specifies the maximal change in output for a given input perturbation.
This simple relation enables building models that are certifiably robust to constrained perturbations in the input space, \ie, those with $L_2$-norm below $\epsilon$.
This is because the resulting output distance $K\cdot \epsilon$ can be calculated efficiently (\ie, in a closed form) without expensive test-time randomization, such as randomized smoothing~\cite{cohen2019certified}, or network bound relaxations, \eg~\cite{ibp,crownibp}.
This way of obtaining certifiable robustness is also directly linked to large margin classifiers, as illustrated in figure \ref{fig:margin}, 
where a training sample free area of radius $\epsilon$ is created around the decision boundary.
The earliest practical examples implementing Lipschitz margins for certified robustness include~\cite{hein2017formal}, which provided first guarantees on deep networks or LMT~\cite{LMT} and GloRo~\cite{gloro} that design Lipschitz margin losses.

\noindent
A limitation with Lipschitz margin classifiers is their decreased performance, as has been shown on standard datasets like CIFAR-10 or TinyImageNet, both in terms of empirical robustness and clean accuracy.
This is emphasized in particular when comparing to approaches such as adversarial training \cite{madry2018towards,carlini2019evaluating} and its variants \cite{zhang2019theoretically,carmon2019unlabeled,wu2020adversarial}.
Thus, recent work proposed specialized architectures~\cite{cisse2017parseval, bcop, anil2019sorting, soc, cpl, xu2022lot, cayley, zhang2021towards, aol, sll, lbdn} to enforce specific Lipschitz constraints, adjusted losses~\cite{LMT, gloro, soc, zhang2021towards, zhang2022boosting, aol} or tighter upper Lipschitz bounds~\cite{lee2020lipschitz, huang2021locallipb, delattre2023efficient} to address these shortcomings.
Despite existing efforts, clean and robust accuracy still lags behind state-of-the-art empirical results.

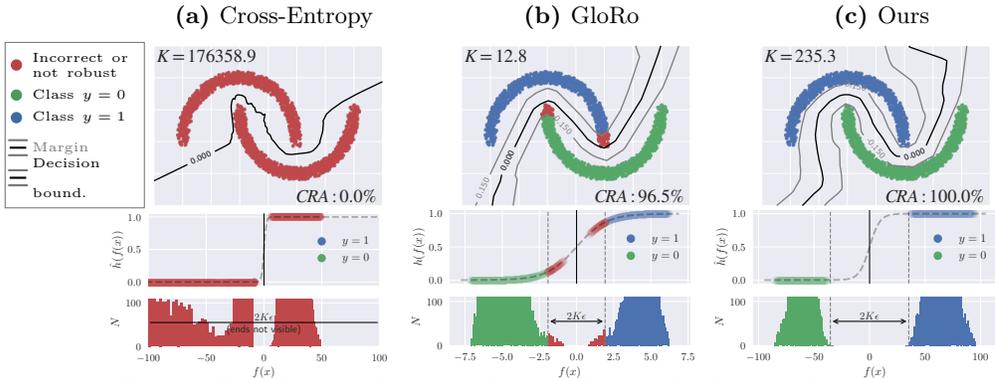
\begin{figure*}
    \centering
    \begin{subfigure}[b]{0.264\textwidth}
        \caption{Cross-Entropy}
        \label{fig:teaser:a}
        \includegraphics[width=\textwidth, trim={1.5cm 1cm 0 0},clip]{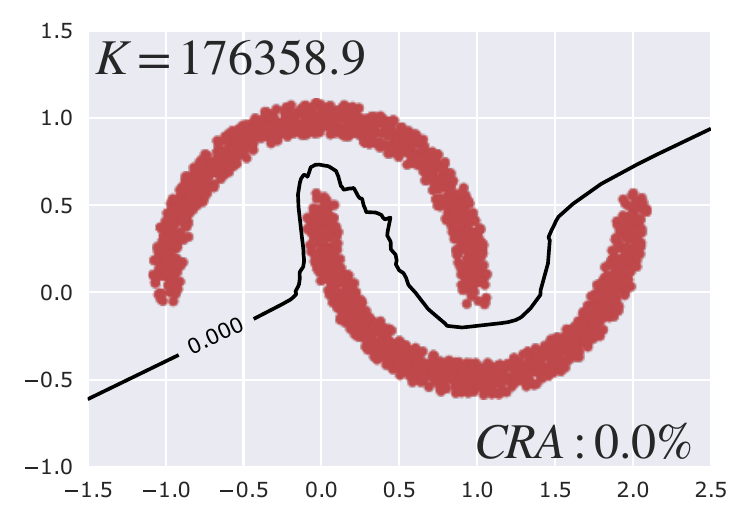}
    \end{subfigure}
    \hspace{0.65cm}
    \begin{subfigure}[b]{0.264\textwidth}
        \caption{GloRo}
        \label{fig:teaser:b}
        \includegraphics[width=\textwidth, trim={1.5cm 1cm 0 0},clip]{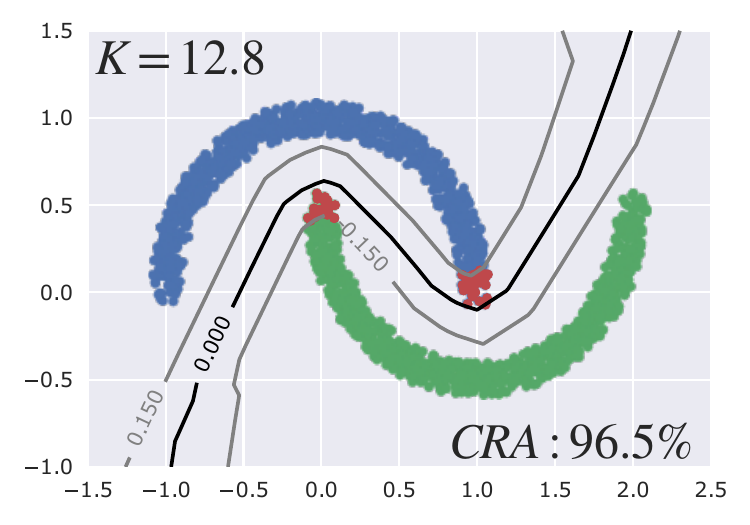}
    \end{subfigure}
    \hspace{0.55cm}
    \begin{subfigure}[b]{0.264\textwidth}
        \caption{Ours}
        \label{fig:teaser:c}
        \includegraphics[width=\textwidth, trim={1.5cm 1cm 0 0},clip]{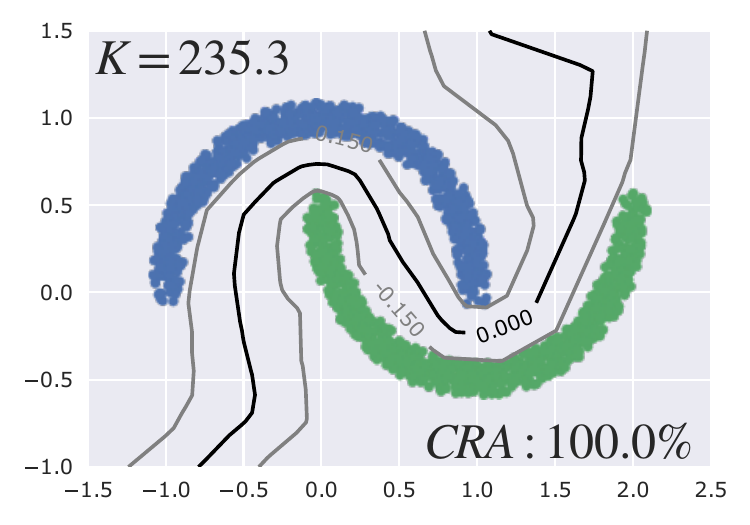}
    \end{subfigure}

    \begin{subfigure}[b]{0.32\textwidth}
        \hspace{-0.4cm}
        \includegraphics[width=\textwidth, trim={0.2cm 0.3cm 0.2 0.3cm},clip]{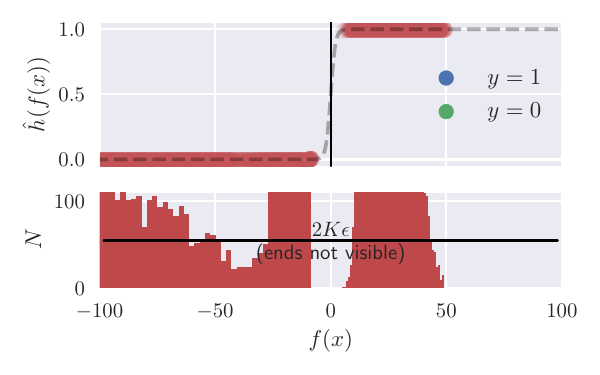}
    \end{subfigure}
    \begin{subfigure}[b]{0.32\textwidth}
        \hspace{-0.4cm}
        \includegraphics[width=\textwidth, trim={0.2cm 0.3cm 0.2cm 0.3cm},clip]{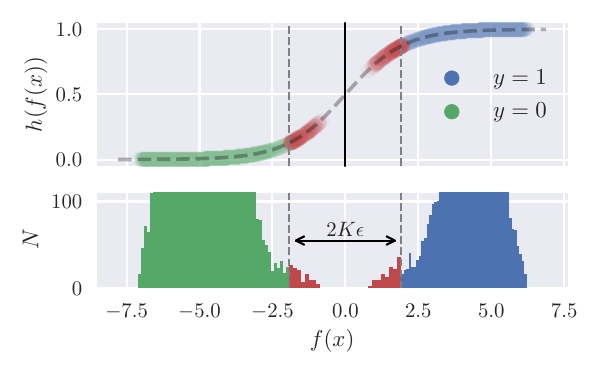}
    \end{subfigure}
    \begin{subfigure}[b]{0.32\textwidth}
        \hspace{-0.4cm}
        \includegraphics[width=\textwidth, trim={0.2cm 0.3cm 0.2cm 0.3cm},clip]{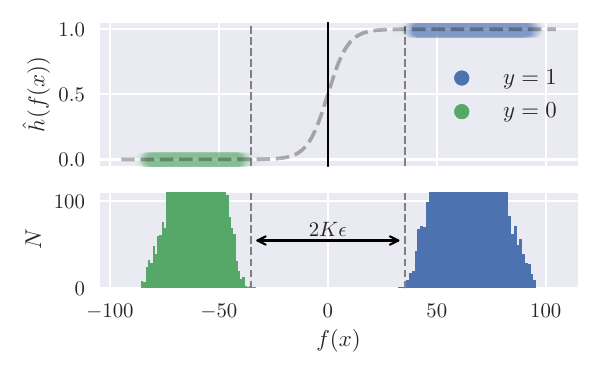}
    \end{subfigure}
    \begin{tikzpicture}[overlay,remember picture,yshift=-0.3cm,xshift=-13.2cm]
        \draw[draw=legendbck,fill=white] (-0.45,2.6) rectangle ++(1.85cm,2.2cm);
        \draw[legendred, fill=legendred] (-0.3,4.5) circle (0.5ex);
        \node[right, text width=1.5cm] at (-0.2,4.6){\tiny Incorrect or};
        \node[right, text width=1.5cm] at (-0.2,4.4){\tiny not robust};
        \draw[legendgreen, fill=legendgreen] (-0.3,4.1) circle (0.5ex);
        \node[right, text width=1.5cm] at (-0.2,4.1){\tiny Class $y=0$};
        \draw[legendblue, fill=legendblue] (-0.3,3.8) circle (0.5ex);
        \node[right, text width=1.5cm] at (-0.2,3.8){\tiny Class $y=1$};
        \draw[draw=legendbck, thick] (-0.41,3.5) -- (-0.15,3.5);
        \draw[draw=black, thick] (-0.39,3.4) -- (-0.17,3.4);
        \draw[draw=legendbck, thick] (-0.41,3.3) -- (-0.15,3.3);
        \node[right, text width=1.5cm] at (-0.2,3.4){\tiny \textcolor{legendbck}{Margin}};
        \draw[draw=legendbck, thick] (-0.41,3.1) -- (-0.15,3.1);
        \draw[draw=black, thick] (-0.39,3.0) -- (-0.17,3.0);
        \draw[draw=legendbck, thick] (-0.41,2.9) -- (-0.15,2.9);
        \node[right, text width=1.5cm] at (-0.2,3.0){\tiny Decision bound.};
    \end{tikzpicture}

    \vspace*{-12px}
    \caption{
    We illustrate how our loss \emph{calibrates} the logistic function of the loss \wrt{} the margin, leading to improved clean and certified robust accuracy (CRA). (a) Cross-Entropy does not guarantee margins, the Lipschitz constant $K$ is large. (b) Uncalibrated margin losses such as GloRo may produce violated margins (gray) and decision boundaries (black) on the two-moons dataset. (c) In contrast, our calibrated loss is better visually and quantitatively for robust and accurate samples. The middle row plots their output probabilities.}
    \label{fig:teaser}
\end{figure*}

We relate this shortcoming to two interlinked properties of used losses:
Firstly, logistic based margin losses, including 
GloRo~\cite{gloro}, SOC~\cite{soc} and CPL~\cite{cpl} do not adjust their output penalties to the output scaling caused by minimization of $K$.
In practice, this results in samples remaining in under-saturated regions of the loss and effectively within margins.
And secondly, we identify that controlling the Lipschitz constant $K$ to be too low, may result in overly smooth decision functions -- impairing clean and robust accuracy.
The result is exemplarily illustrated in figure~\ref{fig:teaser:b} (middle) for GloRo on the two moons dataset.
The induced overly smooth decision surface eventually leads to reduced clean and robust accuracy.
\textbf{Contributions:}
In this work,
we propose \acf{ours}, a loss with improved control over slack and Lipschitz constant $K$ to instrument models with a margin of width $2\epsilon$.
Key to our loss is integrating the scale of the logistic loss with $K$, allowing calibration to the margin.
This calibration endows \ac{ours} with explicit control over slack and $K$, which can improve certified robust accuracy.
As a result, the classifier avoids learning an overly smooth decision boundary while obtaining optimal margin, as illustrated for two moons in figure~\ref{fig:teaser:c}.
On CIFAR-10, CIFAR-100 and Tiny-ImageNet, our method allows greater Lipschitz constants while at the same time having tighter bounds and yielding competitive certified robust accuracies (CRA).
\Ie{} applying \ac{ours} to existing training frameworks consistently improves CRA by multiple percentage points while maintaining or improving clean accuracy.

\section{Related Work}
\myparagraph{Lipschitz classifiers for robustness.}
For completeness, we note that many methods exist for certified robustness.
For an extensive overview we refer the reader to a recent published summary~\cite{li2023sok}.
Since their discovery in \cite{szegedy2014}, many methods for certified robustness have been published \cite{li2023sok}.
Among others, Lipschitz regularization was considered as a potential way to improve robustness against adversarial examples.
While such regularizers could only improve empirical robustness~\cite{cisse2017parseval,jakubovitz2018improving,hoffman2019robust}, local Lipschitz bounds where used to obtain certified robustness in \cite{hein2017formal} at the expense of additional calculations at inference.
The use of global Lipschitz bounds has a key advantage over most other methods: certification at inference is deterministic and cheap to compute.
Alas, it was originally considered intractable due to very loose bounds \cite{hein2017formal,virmaux2018lipschitz}.
More recently, however, a number of methods allow robustness certification based on global Lipschitz bounds, including \cite{bcop, anil2019sorting, LMT, soc, cpl, aol, sll, lbdn, cayley, gloro, lee2020lipschitz, huang2021locallipb, zhang2021towards, zhang2022boosting}.
These improvements can fundamentally be attributed to two independent developments: (i) preventing gradient norm attenuation by specialized non-linearities to preserve classifier complexity~\cite{bcop,anil2019sorting} and (ii) architectural constraints that guarantee a Lipschitz constant of $1$~\cite{bcop, cayley, soc, cpl, xu2022lot, aol, sll, lbdn}.
Additional work investigated the use of tighter bound estimation~\cite{virmaux2018lipschitz,bcop,fazlyab2019efficient,jordan2020exactly,huang2021locallipb,delattre2023efficient}.
Nevertheless, training deep Lipschitz networks with appropriate expressive power remains an open problem \cite{huster2018limitations, rosca2020case}.
We approach this issue using a novel loss that provides more control over the Lipschitz constant which bounds expressiveness~\cite{bartlett2002rademacher}.

\myparagraph{Large margin for deep learning.}
Large margin methods have a long standing history in machine learning to improve generalization of learning algorithms, \eg, see ~\cite{vapnik1982estimation,vapnik1999overview,svm}.
While margin optimization is actively researched in the context of deep learning, as well \cite{sun2014deep,sun2015large,sokolic2017robust,bartlett2017spectrally,elsayed2018large,margin_survey}, state-of-the-art deep networks usually focus on improving accuracy and remain vulnerable to adversarial examples~\cite{madry2018towards,carlini2019evaluating}. This indicates that the obtained margins are generally not sufficient for adversarial robustness.
Moreover, the proposed margin losses often require expensive pairwise sample comparison~\cite{sun2014deep}, normalization of the spectral norm of weights~\cite{bartlett2017spectrally} or linearization of the loss~\cite{elsayed2018large, wei2019improved, ding2019mma}.
Adversarial training can also be viewed as large margin training \cite{ding2019mma}, but does not provide certified robustness. 
This is the focus of our work.

\section{\acf{ours}}

We propose a new margin loss called \ac{ours} that calibrates the loss to the width of the margin -- a property not explicitly accounted for in existing large margin Lipschitz methods.
Specifically, we calibrate the logistic functions at the output (sigmoid and softmax), by
integrating the Lipschitz constant $K$ into the definition of the logistic scale parameter.
This new formulation reveals two properties important for margin training: (i) the scale parameter controls slackness and (ii) slackness influences classifier smoothness.
This slack control allows to trade-off certified robust and clean accuracy.
But more importantly, we find that this slackness determines $K$ of the whole model. This is illustrated on the two moons dataset in figure~\ref{fig:margin} (left) with high slack implying small $K$ and right with low slack implying large $K$.
Given this improved control, we train models with large constants that produce new state-of-the-art robust accuracy scores. 

\subsection{Background}
\label{sec:method:background}

Let $(x, y) \in \mathcal{D_X}$ be a sample-label pair in a dataset in space $\mathcal{X}$, $f$ be a classifier $f: \mathcal{X} \to \mathcal{S}$ where $\mathcal{S} \subseteq \mathbb{R}^N$ is the logit space with $N$ logits and $h$ be a non-linearity like the logistic function or softmax mapping, \eg{} $h: \mathcal{S} \to \mathbb{R}^N$.

\myparagraph{Lipschitz continuity.}
\label{sec:background:lipschitz}
The Lipschitz continuity states that for every function $f$ with bounded first derivative, there exists a Lipschitz constant $K$ that relates the distance between any two points $x_1, x_2$ in input space $\mathcal{X}$ to a distance in output space.
That is, for any input sample distance $\|x_1-x_2\|_p$, the resulting output distance $\|f(x_1)-f(x_2)\|_p$ is at most $K \cdot \|x_1-x_2\|_p$.
Consequently, this inequality enables the construction of losses that measure distances in input space: e.g. the margin width -- and importantly: quick input certification.
\Eg{} assume a classifier with two logits $f_1, f_2$.
An input $x$ is certified robust \wrt{radius} $\epsilon$ when the Lipschitz bounded distance $\epsilon K$ is less than the distance between the two logits: $\epsilon K < |f_1(x)-f_2(x)|$.
Note that $K$ is not required to be small to allow certification.
The inequality also holds when $K$ is large, as long as the distance between logits is greater, as we will see in our experiments in section~\ref{sec:evaluation}.
The exact Lipschitz constant, though, is non-trivial to estimate for highly non-linear functions such as deep networks~\cite{virmaux2018lipschitz}.
Fortunately, it is tractable to calculate an upper bound.
For this, it is sufficient to be able to decompose the classifier $f$, \eg, a (convolutional) neural network, into its $L$ individual layers, i.e. $f = g^{(L)}\circ g^{(L-1)}\circ\dots\circ g^{(1)}$.
Then $K$ is upper bounded by the product of norms~\cite{szegedy2014}, 
{
\setlength{\abovedisplayskip}{0pt}
\setlength{\belowdisplayskip}{2pt}
\begin{equation}
    K \leq \prod_{l=1}^{L} \|g^{(l)}\|_p =: \hat{K}.
    \label{eq:lip_upper_bound}
\end{equation}
}
In general, this bound is extremely loose without any form of Lipschitz regularization\cite{virmaux2018lipschitz}, yet recent work has shown substantially improved tightness, rendering uses in losses tractable (\eg~\cite{gloro, soc, cpl, aol, sll, lbdn, huang2021locallipb}).

\myparagraph{Lipschitz Margin losses.}
To produce certified robust models, we strive to train models with large margins.
This can be achieved by adapting the training objective and specifying a minimal distance $\epsilon$ between any training sample and the decision boundary.
A typical example is the hinge loss formulation in SVMs~\cite{svm}:
{
\setlength{\abovedisplayskip}{4pt}
\setlength{\belowdisplayskip}{2pt}
\begin{align}
\min_{f} \mathbb{E}_{(x,y)\sim\mathcal{D_X}}\left[h\left(\epsilon - yf(x)\right) + \lambda \|f\|^2\right],
\label{eq:svm}
\end{align}
}
where $\epsilon=1$ and $h(\,\cdot\,)=\max\{\,\cdot\,, 0\}$.
$\|f\|$ denotes a generic measure of classifier complexity, which in the linear SVM case is the norm of the weight matrix.
This formulation can be generalized to Lipschitz classifiers,
e.g., by minimizing $\|f\|=K$ or by multiplying $\epsilon$ with its Lipschitz factor $\epsilon K$.
The latter being used in the GloRo loss~\cite{gloro}.
All variants of formulation~\eqref{eq:svm} require strict minimization of $K$ to produce margins.
We find these types of losses can be improved when $h$ belongs to the logistic family -- as sigmoid and softmax are.
Since logistic functions assume a fixed output distribution, we observe that minimizing $K$ can leave samples too close to the decision boundary
We present exemplary evidence in figure~\ref{fig:teaser:b} in which red samples remain within the margins.
This is specifically true for GloRo, but also for methods that hard constrain $K$ (e.g. \cite{soc, cpl, xu2022lot}). 
In the following, we address these issues with \ac{ours}.

\subsection{Binary \ac{ours}}
\label{sec:method:ours_bce}

We base our construction of \ac{ours} on the general margin formulation in equation~\eqref{eq:svm}.
Key is calibrating the output non-linearity $h$ to the margin.
In the binary case, it is common practice in deep learning to set $h$ in equation~\eqref{eq:svm} to a logistic function $h(x)={[1+\exp\{-x/\sigma\}]^{-1}}$ with fixed $\sigma = 1$ and minimize the binary cross-entropy loss.
Yet the underlying logistic distribution assumes a fixed distribution width $\sigma=1$, which can be detrimental for training Lipschitz classifiers.
To demonstrate the limitation of this assumption, we look at figure~\ref{fig:teaser} illustrating margin training on the two-moons dataset.
Figure~\ref{fig:teaser:a} illustrates vanilla cross-entropy and no Lipschitz regularization.
The decision boundary attains an irregular shape without margin. 
The Lipschitz constant is very high, the output values attain high probabilities since they are pushed to the tails of the distribution (see $p=h(f(x))$ in middle row).
In contrast, a Lipschitz margin loss (GloRo~\cite{gloro}, fig.~\ref{fig:teaser:b}) produces a margin, yet non-robust samples (red) remain within margin boundaries.
This is a consequence of minimizing $K$, which limits the spread of the distribution. 
Under this condition, the assumption $\sigma=1$ is inefficient. 
Additionally, since $\sigma$ of the logistic is fixed, the final probability $p$ at the margin $\pm \epsilon$ can only be determined post-hoc, after the Lipschitz constant $K$ finds a minimum.
We can be more efficient about this process by calibrating the width to $K\epsilon$ at each training step and requiring $p$ at $\pm \epsilon$ to be a specific value.
The result is shown in figure~\ref{fig:teaser:c} which produces the desired margin with no errors.
Our proposed loss is defined as follows.

\begin{figure}[t]
    \centering
    \vspace{-0.5em}
    \begin{subfigure}[t]{0.25\columnwidth}
        \centering
        \includegraphics[width=\textwidth, trim={0.35cm 0 0 0.4cm}, clip]{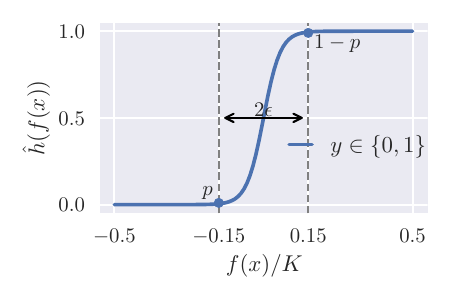}
    \end{subfigure}
    \begin{subfigure}[t]{0.25\columnwidth}
        \centering
        \includegraphics[width=\textwidth, trim={0.35cm 0 0 0.4cm}, clip]{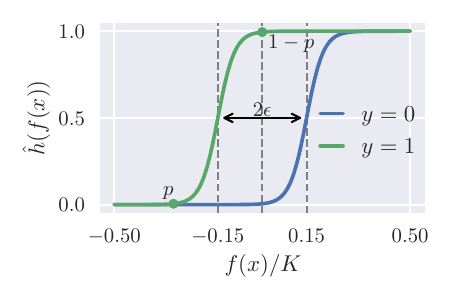}
    \end{subfigure}
    \begin{subfigure}[t]{0.25\columnwidth}
        \centering
        \includegraphics[width=\textwidth, trim={0.4cm 0 0.2cm 0.2cm},clip]{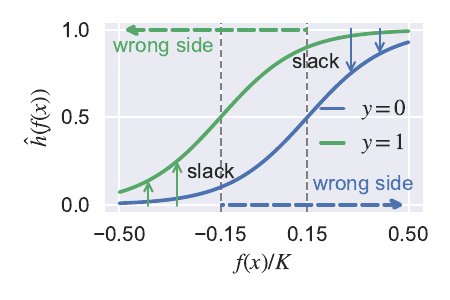}
    \end{subfigure}
    \vspace{-1.5em}
    \caption{
    Calibrated logistic function as key to our \ac{ours} loss by parameterizing the scale parameter $\sigma$ to margin width $2\epsilon$. Left: without margin offset; middle: with margin offset $\epsilon$; right: slack in \ac{ours} is governed by parameter $p$. That is to increase slack, we increase $p$ which decreases loss for samples on the wrong margin side (indicated by arrows).
    }
    \label{fig:calibrated_logistic_basic}
    \label{fig:calibrated_logistic}
    \label{fig:logistic_slack}
    \vspace{-1em}
\end{figure}

\noindent
\myparagraph{Definition 1 - Binary \ac{ours}.} \textit{Let $y\in\{-1,1\}$, 
and $\mathcal{L}$ be the binary cross entropy loss $\mathcal{L}(y,h(f(x)))=-\log h(f(x)) - (1-y)\log (1-h(f(x)))$}\footnote{Transformation to $y\in\{0,1\}$ is omitted for readability.}.
We propose the objective:
{
\setlength{\abovedisplayskip}{0pt}
\setlength{\belowdisplayskip}{2pt}
\begin{align}
    \min_{f}&\,\, \mathbb{E}_{(x,y)\sim\mathcal{D_X}}\left[\mathcal{L}(y, \hat{h}(f(x);y))+ \lambda K^2\right]\label{eq:bce_optimization}\\
    \text{with }\, &\hat{h}(f(x);y) = h\left(-\frac{y\epsilon}{\sigma_\epsilon(p)} + \frac{1}{\sigma_{\epsilon}(p)}\frac{f(x)}{K}\right)\label{eq:bce_full}\\
    \text{\text{and} }\, &\sigma_{\epsilon}(p) = \frac{2\epsilon}{h^{-1}(1-p) - h^{-1}(p)},
    \label{eq:sigma}
\end{align}
}
where $\hat{h}$ is our \textit{calibrated logistic} and $h^{-1}$ is the inverse cdf: $h^{-1}(p)=\log(\nicefrac{p}{1-p})$.
Our proposal follows from calibrating $\sigma$ to $2\epsilon$.
For its realization we only require $4$ values to uniquely determine $\sigma$: (i) the two positions of the margin boundaries $\pm \epsilon$ and (ii) the two probabilities $p_{-\epsilon}, p_\epsilon$ that the logistic distribution should attain at these positions.
We illustrate these $4$ values in figure~\ref{fig:calibrated_logistic_basic},
which displays an already calibrated logistic function to $\epsilon=0.15$ (vertical lines) and $p_{-\epsilon}=p=10^{-3}$ and $p_{\epsilon}=1-p=1-10^{-3}$.
The theoretical derivation follows from the logistic distribution, which we discuss in the supplement sec.~A.
We denote the calibrated scale as $\sigma_\epsilon(p)$ as it depends on $\epsilon$ and a probability $p$ (equation~\eqref{eq:sigma}).
So far, this does not express the integration of the Lipschitz constant $K$.
Recall that the input distance $2\epsilon$ can be related to the corresponding output distance via $K$, \ie{} $2K\epsilon$.
We consequently acquire the Lipschitz calibrated version shown in figure~\ref{fig:calibrated_logistic_basic}: $\nicefrac{f(x)}{\sigma_\epsilon(p) K}$.
Next, we integrate hard margin offsets $\pm \epsilon$ as in formulation~\eqref{eq:svm} to maximize the penalty on samples within the margin.
To integrate, we add $\pm \epsilon K$, depending on the class sign: $[-yK\epsilon + f(x)]/[\sigma_\epsilon(p) K]$.
This places the probability on the margin to the worst case value $0.5$.
Note that this does not invalidate our calibration: the offset is a tool to improve margin training and is removed during inference.
\ac{ours} is easy to generalize to multinomial classification utilizing softmax.
We show its implementation in the supplement, sec.~A.

\subsection{Discussion}
\label{sec:method:discussion}

\ac{ours} is derived from joining the definition of $\sigma$ with the Lipschitz inequality while adjusting for the margin distance $2\epsilon$.
By utilizing the Lipschitz constant, this scale parameter can be calibrated to the margin width as if measured in input space.
This is feasible because of the normalization by the Lipschitz constant $\nicefrac{f(x)}{K}$ in equations~\eqref{eq:bce_full}.
This normalization has two additional ramifications.
First, it decouples the classifier constant from the loss.
$K$ can attain any value, while \ac{ours} ensures calibration.
And secondly, the Lipschitz constant of the whole model $h(f(x))$ is solely dependent on $\sigma_{\epsilon}(p)$. 
That is $\khf = \kh = \nicefrac{1}{\sigma_{\epsilon}(p)}$.
Below, we discuss the implications of \ac{ours} with respect to the tightness of Lipschitz bounds, the interpretation of $\sigma$ as allowed slack and the classifier's complexity.

\myparagraph{Tightness.} 
Tightness dictates the utility of the used Lipschitz bound.
To measure, we can estimate a na\"ive lower bound to $K$ by finding the pair of training samples $x_1,x_2$ that maximizes the quotient $\nicefrac{\|f(x_1)-f(x_2)\|}{\|x_1-x_2\|}$.
Since the input sample distances remain fixed, tightness can only be increased by increasing output distances \textit{and} bounding $K$ -- conceptually bringing the two quantities closer together.
Recall that typical losses assume a fixed output distribution $\sigma=1$.
The capacity to push values into the saturated area of the loss is thereby limited.
Tightness is thereby mainly achieved by minimizing $K$ because output distances cannot be increased significantly.
\ac{ours} instead, does not minimize $K$ but bounds it via normalization.
Since our loss is calibrated to $K\epsilon$, we can achieve increased tightness by only maximizing output distances.
With the difference of increasing output distances much farther than possible with other margin losses (an example is illustrated in figure~\ref{fig:teaser} (compare x-axis of (b) and (c)).
We find this to converge faster (see figure~D1 in supplement) and achieve better robust accuracies than related work, as we present in section~\ref{sec:evaluation}.
However, since increasing output distances has a growth influence on $K$, we find it necessary to put slight regularization pressure on $K$ via factor $\lambda$, as stated on the RHS of equation~\eqref{eq:bce_optimization}. 

\myparagraph{Slack.}
\label{sec:method:slack}
Since the logistic function has a smooth output transition from $0$ to $1$, values on the wrong margin side can be interpreted as slackness.
We illustrate slack in figure~\ref{fig:logistic_slack} which shows a calibrated logistic as in figure~\ref{fig:calibrated_logistic} and indicates wrong margin sides for each class by dashed arrows.
Consider class $y=1$ in green.
If a sample falls on the wrong margin side (anywhere along the green arrow at the top) the probability decreases and consequently the loss increases (\ie, negative log-likelihood $\lim_{p\to 0}-\log(p)=\infty$).
The governing factor for slack in our loss is the probability $p$ in $\sigma_{\epsilon}(p)$ of equation~\ref{eq:bce_full}, indicated by the green vertical arrows, which implicitly defines the loss magnitude for samples on the wrong margin side.
Note that without calibration, we find the logistic $\sigma$ or temperature in softmax to have no slack effect.
We provide empirical evidence in the supplement, sec.~D.

\begin{figure*}[t]
    \centering
    \vspace*{-0.5em}
    \begin{minipage}[t]{0.014\textwidth}
        \vspace{3em}
        \rotatebox[origin=c]{90}{Ours}
    \end{minipage}
    \begin{subfigure}[t]{0.2\textwidth}
        \centering
        \caption{$2\epsilon=0.2$}
        \label{fig:moons_a}
        \includegraphics[width=\textwidth]{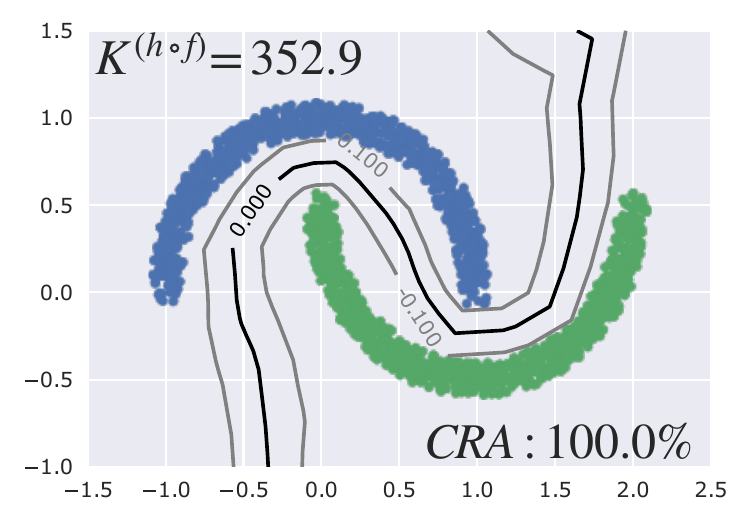}
    \end{subfigure}
    \hfill
    \begin{subfigure}[t]{0.2\textwidth}
        \centering
        \vspace*{-1em}
        \caption{$2\epsilon=0.3$\\(True margin)}
        \label{fig:moons_b}
        \includegraphics[width=\textwidth]{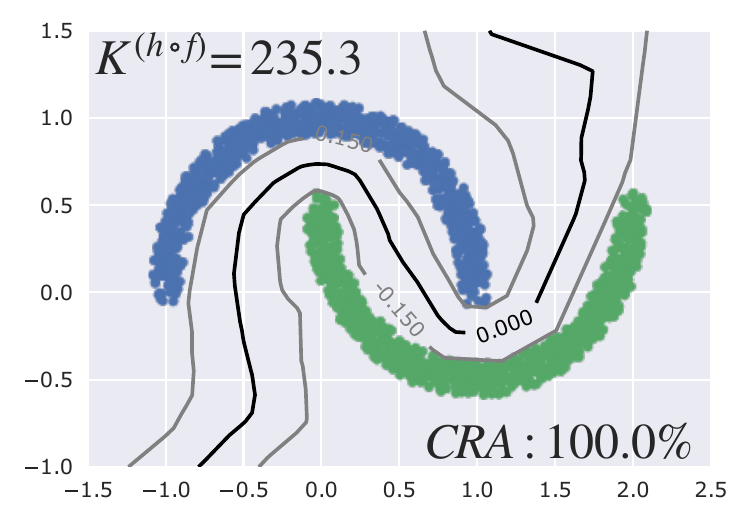}
    \end{subfigure}
    \hfill
    \begin{subfigure}[t]{0.2\textwidth}
        \centering
        \caption{$2\epsilon=0.4$}
        \label{fig:moons_c}
        \includegraphics[width=\textwidth]{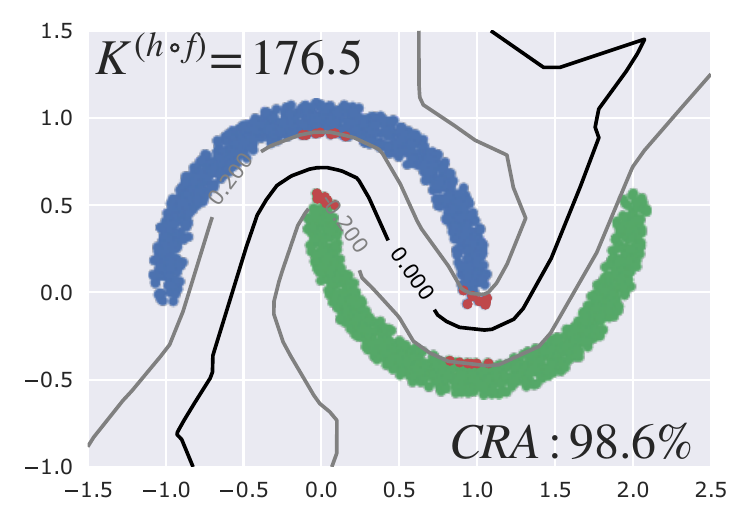}
    \end{subfigure}
    \hfill
    \begin{subfigure}[t]{0.2\textwidth}
        \centering
        \caption{$2\epsilon=0.5$}
        \label{fig:moons_d}
        \includegraphics[width=\textwidth]{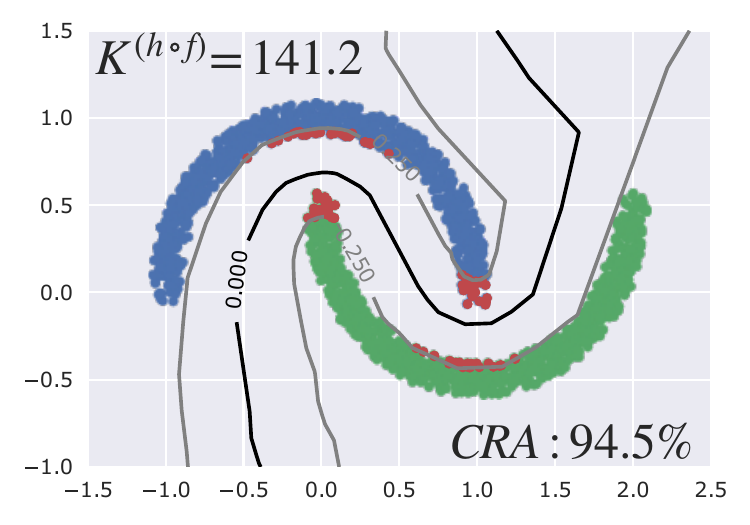}
    \end{subfigure}
    \hfill
    
    \begin{minipage}[t]{0.014\textwidth}
        \vspace{0.5em}
        \rotatebox[origin=c]{90}{GloRo}
    \end{minipage}
    \begin{subfigure}[t]{0.2\textwidth}
        \centering
        \vspace{-0.8em}
        \includegraphics[width=\textwidth]{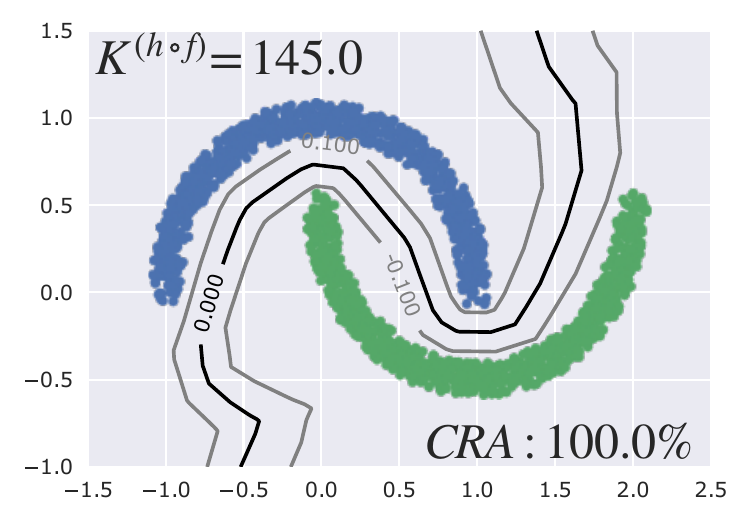}
    \end{subfigure}
    \hfill
    \begin{subfigure}[t]{0.2\textwidth}
        \centering
        \vspace{-0.8em}
        \includegraphics[width=\textwidth]{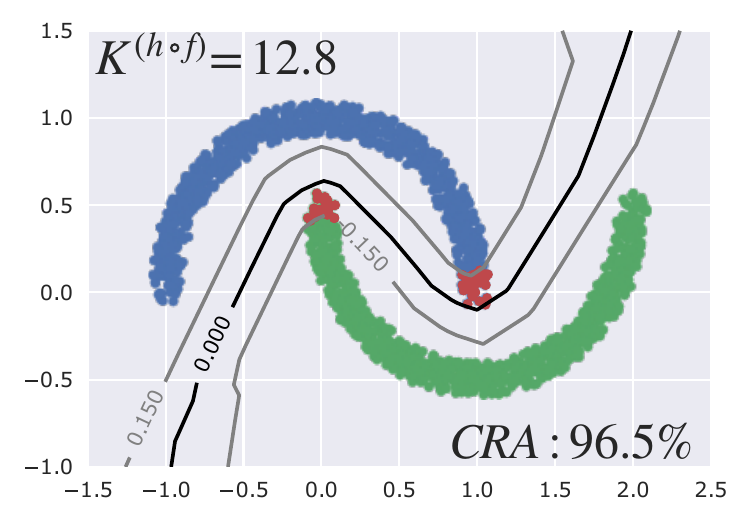}
    \end{subfigure}
    \hfill
    \begin{subfigure}[t]{0.2\textwidth}
        \centering
        \vspace{-0.8em}
        \includegraphics[width=\textwidth]{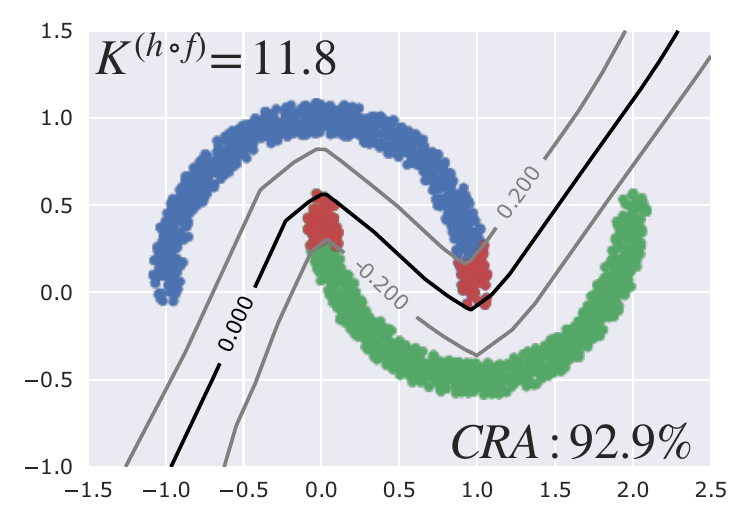}
    \end{subfigure}
    \hfill
    \begin{subfigure}[t]{0.2\textwidth}
        \centering
        \vspace{-0.8em}
        \includegraphics[width=\textwidth]{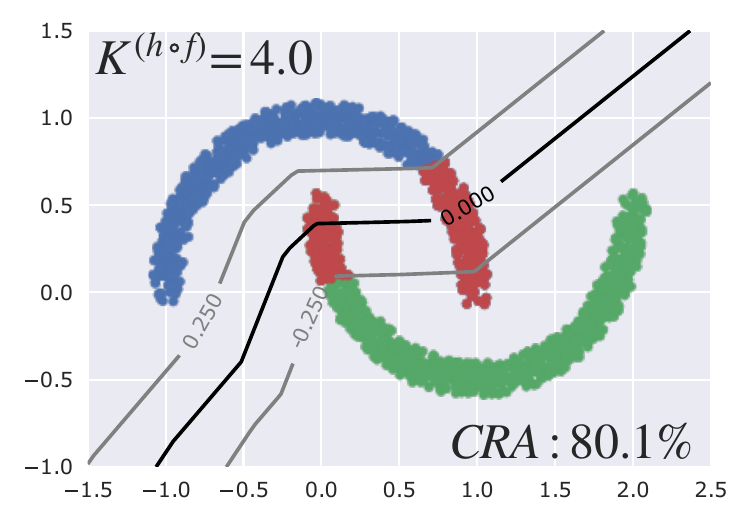}
    \end{subfigure}
    \hfill
    
    \vspace{-0.5em}
    \caption{
        Our \ac{ours} (top row) vs margin loss GloRo (bottom) on two-moons.
        \textcolor[rgb]{0.70, 0.31, 0.30}{\bf Red points} denote incorrect or non-robust samples.
        GloRos formulation leads to consistently smaller Lipschitz constant $\khf$ restricting the classifier complexity.
        Already for the true margin $2\epsilon=0.3$ it fits inefficiently.
        For ``too-large'' margins, GloRo eventually degenerates to a failure case with a near linear decision boundary.
        In contrast, our \ac{ours} produces a $100\%$ accurate and robust model for the true margin and retains sensible margins for increased $\epsilon$.}
    \label{fig:moons}
\end{figure*}

\myparagraph{Classifier complexity.}
\cite{bartlett2002rademacher} derive complexity bounds for a $2$-layered neural network (theorem 18).
\Ie{}, given $2$ layers with weight norm $B$ and $1$ and non-linearity with Lipschitz constant $L$, the Gaussian complexity of the model is bounded by their product $K = B\cdot L \cdot 1$.
This directly applies to hard-constrained $K$=$1$ models, 
but we find it applies to soft-constrained models as well.
Effectively, this bound has implications for training Lipschitz classifiers: if $K$ is too small, it restricts the models ability to fit the data.
While theory provides only an upper bound, we find empirical evidence as support.
That is, clean and certified robust accuracy can be improved when $L$ of the loss is adjusted -- and thus $K$.
We find strongest evidence on two-moons (figure~\ref{fig:moons}, bottom row), and consistent support on image data (section~\ref{sec:evaluation}).
With decreasing $K$ (top), the model becomes overly smooth, losing performance.
\ac{ours} offers direct control over $K$, simply by adjusting slack.
That is, $K$ is inversely proportional to the slack $p$: $\khf=\nicefrac{1}{\sigma_\epsilon (p)}$.
Consequently, for a fixed $\epsilon$, slack bounds the complexity of the model.
Models that can separate the data well require little slack, implying a larger $\khf$.
The next section discusses this in more detail.

\section{Evaluation}
\label{sec:evaluation}

\ac{ours} offers increased control over slack and classifier complexity, discussed in section~\ref{sec:method:discussion}.
In this section, we present empirical evidence for these claims.
We show that decreasing slack leads to models with less smooth decision boundaries, resulting in higher clean and certified robustness.
To this end, we present results on the synthetic two-moons dataset by visualizing the produced margins and
discuss the application on natural images: CIFAR-10, CIFAR-100~\cite{krizhevsky2009learning} and Tiny-ImageNet~\cite{le2015tiny}.
Implementation of our method involves computing the upper Lipschitz bound.
We measure the width of the margin with the $L_2$ distance.
Consequently, we implement equation~\eqref{eq:lip_upper_bound} with the product of spectral norms~\cite{szegedy2014} and calculate them by performing power iterations.
Our training strategy follows the respective method we compare with.
That is, we reuse published code (where available) but use our loss.
To evaluate, we measure certified robust accuracies (CRA) for three margin widths ($36/255$, $72/255$, $108/255$) and Lipschitz bound tightness.
CRA represents the average number of accurately classified samples that are also robust (fall outside the margin).
The latter is the fraction between empirical lower bound and upper bound.
All training and metric details are listed in the supplement, sec.~B and~C. 
Our code is made publicly available at \href{https://github.com/mlosch/CLL}{github.com/mlosch/CLL}.

\subsection{Two-moons dataset}
We start with an analysis on two-moons in order to visualize learned decision boundaries and investigate the connection to the Lipschitz constant.
using \textit{uniform} sampled noise of radius $0.1$ around each sample.
This results in a true-margin of exactly $2\epsilon=0.3$.
In figure~\ref{fig:moons}, we exemplarily compare \ac{ours} with GloRo, training 7-layered networks for different target margin widths (columns).
Individual plots display the decision boundary (black line) and the Lipschitz margin (gray lines).
CRAs and Lipschitz constants $\khf$ are reported in the corners, non-robust or misclassified training samples are marked red.
GloRo already loses CRA at the true margin $2\epsilon$ and the decision boundary becomes very smooth with decreasing $\khf$.
In contrast, \ac{ours} retains $100\%$ CRA for the true margin and only slowly loses CRA beyond.
The key is in the control over slack and hence $\khf$ -- we set $p=10^{-12}$ in equation~\ref{eq:bce_full}.
Our decision functions do not become overly smooth.

\subsection{Image datasets}

We continue our discussion on CIFAR-10, CIFAR-100 and Tiny-ImageNet, evaluating multiple architectures:
On CIFAR-10/100, we evaluate \textit{6C2F}\cite{lee2020lipschitz}, \textit{4C3F}\cite{wong2018scaling}, \textit{LipConv}\cite{soc} and \textit{XL}\cite{cpl, sll}.
On Tiny-ImageNet, we consider \textit{8C2F}\cite{lee2020lipschitz}, \textit{LipConv}, XL and LBDN\cite{lbdn}.
We report Tiny-ImageNet results in table~\ref{tbl:results_tinyimagenet} and CIFAR results in table~\ref{tbl:results}, considering CRA for three different margin widths and clean accuracy.
Hereby, all values produced with \ac{ours} are averages over $9$ runs with different random seeds.
Standard deviations are reported in the supplement, sec~D.
Additionally, we report tightness and Lipschitz constants of both the classifier $f$, as well as the composition with the loss $h\circ f$ -- stating the effective complexity of the model.
$\maxkf$ and $\maxkhf$ state the largest constant between pairs of classes, e.g. $\maxkf=\max_{i,j} \hat{K}^\spow{f}_{i,j}$.
Hereby, data scaling factors (e.g. normalization) influence $\khf$.
\Eg{} a normalization factor of $5$, increases $\khf$ by the same factor.

\begin{table*}[t]
\centering
\vspace*{-1em}
\caption{
Results on Tiny-ImageNet on clean and certified robust accuracy (CRA) for six different methods using Lipschitz bounds $\maxkhf$.
Applying \ac{ours} to existing methods consistently improves certified robust accuracy (CRA).
Architectures evaluated: \textit{8C2F}\cite{lee2020lipschitz}, \textit{LipConv}\cite{soc}, \textit{XL}\cite{cpl, sll} and \textit{Sandwich}\cite{lbdn} 
$\top$ indicates model is additionally trained with TRADES-loss\cite{zhang2019theoretically}. 
$\dagger$-flagged numbers are reproduced values with our own code. 
\ac{ours} numbers are averaged over $9$ runs. 
}
\label{tbl:results_tinyimagenet}
\resizebox{\linewidth}{!}{
\begin{tabular}{cc|@{}llccccccc@{}|}
\cline{3-11} 
&& \multicolumn{1}{l|}{Method} & \multicolumn{1}{l|}{Model} & Clean (\%) & CRA $\frac{36}{255}$ (\%) & CRA $\frac{72}{255}$ (\%) & \multicolumn{1}{c|}{CRA $\frac{108}{255}$(\%)} & $\maxkf$ & $\maxkhf$ & Tightness (\%) \\
\midrule
\multirow{11}{2mm}{\rotatebox[origin=c]{90}{\textbf{Tiny-ImageNet}}} &
&
\multicolumn{1}{l|}{GloRo\cite{gloro}} & \multicolumn{1}{l|}{8C2F$^\top$} & 35.5 & 22.4 & - & \multicolumn{1}{c|}{-} & \multicolumn{2}{c}{12.5} & 47 \\
&& \multicolumn{1}{l|}{} & \multicolumn{1}{l|}{8C2F$^\top \dagger$} & 39.5 & 23.9 & 14.3 & \multicolumn{1}{c|}{9.0} & \multicolumn{2}{c}{3.9} & 53 \\ 
\cline{3-11}
&& \multicolumn{1}{l|}{Local-Lip-B\cite{huang2021locallipb}} & \multicolumn{1}{l|}{8C2F} & 36.9 & 23.4 & 12.7 & \multicolumn{1}{c|}{6.1} & - & - & - \\ \cline{3-11}
&& \multicolumn{1}{l|}{Ours $\epsilon=0.5, p=0.01$} & \multicolumn{1}{l|}{8C2F} & \textbf{39.8 (+0.3)} & \textbf{25.9 (+2.0)} & \textbf{16.5 (+2.2)} & \multicolumn{1}{c|}{\textbf{10.7 (+1.7)}} & 288.3 & 10.6 & \textbf{64 (+11)} \\  
\cmidrule{3-11}\morecmidrules\cmidrule{3-11}
&& \multicolumn{1}{l|}{SOC\cite{soc}} & \multicolumn{1}{l|}{LipConv-10} & 32.1 & 21.5 & 12.4 & \multicolumn{1}{c|}{7.5} & \multicolumn{2}{c}{6.4} & \textbf{85} \\
&& \multicolumn{1}{l|}{} & \multicolumn{1}{l|}{LipConv-20} & 31.7 & 21.0 & 12.9 & \multicolumn{1}{c|}{7.5} & \multicolumn{2}{c}{6.4} & 81 \\ 
\cline{3-11}
&& \multicolumn{1}{l|}{Ours} & \multicolumn{1}{l|}{LipConv-20} & \textbf{32.6 (+0.5)} & \textbf{26.0 (+3.5)} & \textbf{20.2 (+7.3)} & \multicolumn{1}{c|}{\textbf{15.5 (+8.0)}} & 4.9 & 11.6 & \textbf{84 (+3)}\\ 
\cmidrule{3-11}\morecmidrules\cmidrule{3-11}
&& \multicolumn{1}{l|}{SLL\cite{sll}} & \multicolumn{1}{l|}{XL} & 32.1 & 23.2 & 16.8 & \multicolumn{1}{c|}{12.0} & - & - & -\\\cmidrule{3-11}
&& \multicolumn{1}{l|}{LBDN\cite{lbdn}} & \multicolumn{1}{l|}{Sandwich} & 33.4 & 24.7 & 18.1 & \multicolumn{1}{c|}{13.4} & - & - & - \\\cmidrule{3-11}
&& \multicolumn{1}{l|}{Ours $\epsilon=1.0, p=0.025$} & \multicolumn{1}{l|}{8C2F} & \textbf{33.5 (+0.1)} & \textbf{25.3 (+0.6)} & \textbf{19.0 (+0.9)} & \multicolumn{1}{c|}{\textbf{13.8 (+0.4)}} & 24.4 & 4.4 & 73 \\
\bottomrule
\end{tabular}
}
\vspace{-1em}
\end{table*}

\begin{table*}[t]
\centering
\vspace*{-1em}
\caption{
Continuation of table~\ref{tbl:results_tinyimagenet} but on CIFAR-10 and CIFAR-100.
Additional architectures evaluated:  \textit{4C3F}\cite{wong2018scaling}, \textit{6C2F}\cite{lee2020lipschitz}.
Local-Lip-B tightness is estimated by reading values off of figure 2a~\cite{huang2021locallipb}. 
}
\label{tbl:results}
\resizebox{\linewidth}{!}{
\begin{tabular}{cc|@{}llccccccc@{}|}
\cline{3-11} 
&& \multicolumn{1}{l|}{Method} & \multicolumn{1}{l|}{Model} & Clean (\%) & CRA $\frac{36}{255}$ (\%) & CRA $\frac{72}{255}$ (\%) & \multicolumn{1}{c|}{CRA $\frac{108}{255}$(\%)} & $\maxkf$ & $\maxkhf$ & Tightness (\%) \\
\midrule
\multirow{13}{2mm}{\rotatebox[origin=c]{90}{\textbf{CIFAR-10}}} 
&& \multicolumn{1}{l|}{Local-Lip-B\cite{huang2021locallipb}} & \multicolumn{1}{l|}{6C2F} & 77.4 & 60.7 & - & \multicolumn{1}{c|}{-} & \multicolumn{2}{c}{7.5} & $\approx$80 \\ 
\cline{3-11}
&& \multicolumn{1}{l|}{\multirow{2}{*}{GloRo\cite{gloro}}} & \multicolumn{1}{l|}{6C2F} & 77.0 & 58.4 & - & \multicolumn{1}{c|}{-} & \multicolumn{2}{c}{15.8} & 70 \\
&& \multicolumn{1}{l|}{} & \multicolumn{1}{l|}{4C3F$\dagger$} & 77.4 & 59.6 & 40.8 & \multicolumn{1}{c|}{24.8} & \multicolumn{2}{c}{7.4} & 71 \\ 
\cline{3-11}
&& \multicolumn{1}{l|}{Ours} & \multicolumn{1}{l|}{6C2F} & \textbf{77.6 (+0.6)} & 61.3 (+2.9) & 43.5 & \multicolumn{1}{c|}{27.7} & 709.3 & 11.6 & 78 (+8) \\
&& \multicolumn{1}{l|}{} & \multicolumn{1}{l|}{4C3F} & 77.6 (+0.2) & \textbf{61.4 (+1.8)} & \textbf{44.2 (+3.4)} & \multicolumn{1}{c|}{\textbf{29.1 (+4.3)}} & 72.1 & 11.6 & \textbf{80 (+9)} \\ 
\cmidrule{3-11}\morecmidrules\cmidrule{3-11}
&& \multicolumn{1}{l|}{SOC\cite{soc}} & \multicolumn{1}{l|}{LipConv-20} & 76.3 & 62.6 & 48.7 & \multicolumn{1}{c|}{\textbf{36.0}} & \multicolumn{2}{c}{5.6} & 86 \\
\cline{3-11}
&& \multicolumn{1}{l|}{Ours} & \multicolumn{1}{l|}{LipConv-20} & \textbf{77.4 (+1.1)} & \textbf{64.2 (+1.6)} & \textbf{49.5 (+0.8)} & \multicolumn{1}{c|}{\textbf{36.7 (+0.7)}} & 35.8 & 14.7 & 86 \\
\cmidrule{3-11}\morecmidrules\cmidrule{3-11}
&& \multicolumn{1}{l|}{CPL\cite{cpl}} & \multicolumn{1}{l|}{XL} & 78.5 & 64.4 & 48.0 & \multicolumn{1}{c|}{33.0} & \multicolumn{2}{c}{$\sqrt{2}$} & 78 \\ 
\cline{3-11}
&& \multicolumn{1}{l|}{Ours} & \multicolumn{1}{l|}{XL} & \textbf{78.8 (+0.3)} & \textbf{65.9 (+1.5)} & \textbf{51.6 (+3.6)} & \multicolumn{1}{c|}{\textbf{38.1 (+5.1)}} & 34.6 & 11.8 & \textbf{80 (+2)}  \\
\cmidrule{3-11}\morecmidrules\cmidrule{3-11}
&& \multicolumn{1}{l|}{SLL\cite{sll}} & \multicolumn{1}{l|}{XL} & \textbf{73.3} & \textbf{65.8} & \textbf{58.4} & \multicolumn{1}{c|}{\textbf{51.3}} & $\sqrt{2}$ & 6.7 & 88  \\ 
\cline{3-11}
&& \multicolumn{1}{l|}{Ours} & \multicolumn{1}{l|}{XL} & 73.0 & 65.5 & 57.8 & \multicolumn{1}{c|}{51.0} & 59.4 & 10.6 & 88.0  \\ 
\bottomrule
\multirow{7}{2mm}{\rotatebox[origin=c]{90}{\textbf{CIFAR-100}}} &
&
\multicolumn{1}{l|}{SOC\cite{soc}} & \multicolumn{1}{l|}{LipConv-20} & 47.8 & 34.8 & 23.7 & \multicolumn{1}{c|}{15.8} & \multicolumn{2}{c}{6.5} & \textbf{85 (+1)} \\
\cline{3-11}
&& \multicolumn{1}{l|}{Ours} & \multicolumn{1}{l|}{LipConv-20} & \textbf{48.2 (+0.4)} & \textbf{35.1 (+0.3)} & \textbf{25.3 (+1.6)} & \multicolumn{1}{c|}{\textbf{18.3 (+2.5)}} & 45.4 & 9.2 & 84 \\
\cmidrule{3-11}\morecmidrules\cmidrule{3-11}
&& \multicolumn{1}{l|}{CPL\cite{cpl}} & \multicolumn{1}{l|}{XL} & 47.8 & 33.4 & 20.9 & \multicolumn{1}{c|}{12.6} & \multicolumn{2}{c}{1.6} & 74 \\ 
\cline{3-11}
&& \multicolumn{1}{l|}{Ours} & \multicolumn{1}{l|}{XL} & \textbf{47.9 (+0.1)} & \textbf{36.3 (+2.9)} & \textbf{28.1 (+7.2)} & \multicolumn{1}{c|}{\textbf{21.5 (+8.9)}} & 42.0 & 7.6 & \textbf{79 (+5)}  \\
\cmidrule{3-11}\morecmidrules\cmidrule{3-11}
&& \multicolumn{1}{l|}{SLL\cite{sll}} & \multicolumn{1}{l|}{XL} & 46.5 & 36.5 & 29.0 & \multicolumn{1}{c|}{23.3} & 1.5 & 6.0 & \textbf{81}  \\ 
\cmidrule{3-11}
&& \multicolumn{1}{l|}{Ours} & \multicolumn{1}{l|}{XL} & \textbf{46.9 (+0.4)} & \textbf{36.6 (+0.1)} & 29.0 & \multicolumn{1}{c|}{\textbf{23.4 (+0.1)}} & 1.3 & 6.5 & 80  \\ 
\bottomrule
\end{tabular}
}
\vspace{-1em}
\end{table*}

\myparagraph{Certified robust accuracy.} 
For fair comparison, we adjust $\epsilon$ and $p$ of \ac{ours} to match or outperform clean accuracy of the respective baseline (details in supplement, sec.~D).
First, we consider Tiny-ImageNet (table~\ref{tbl:results_tinyimagenet}).
\textit{8C2F} is used to compare GloRo, \textit{Local-Lip-B} and \ac{ours}, \textit{LipConv} is used to compare \textit{SOC} and \ac{ours} and \textit{SLL} on \textit{XL} and \textit{LBDN} on \textit{Sandwich} is compared to \ac{ours} on \textit{8C2F}.
Here, GloRo on \textit{8C2F} achieves $23.9\%$ CRA for $\epsilon=36/255$ and $39.5\%$ clean accuracy.
Trained with \ac{ours}, we achieve a substantial increase for the same margin of $25.9\% (+2.0)$ while improving clean accuracy $39.8\% (+0.3\%)$.
We note that GloRo additionally uses the TRADES-loss~\cite{zhang2019theoretically} on \textit{8C2F} to trade-off clean for robust accuracy.
\ac{ours} simplifies this trade-off control via slack parameter $p$, see section \ref{sec:evaluation:analysis}.
Regarding \textit{SOC} on \textit{LipConv}, we find $10$ layers to perform slightly better than $20$ (training setup in supplement, sec~B).
This is in line with the observation in~\cite{soc}: adding more layers can lead to degrading performance.
\ac{ours}, in contrast, clearly outperforms \textit{SOC} on $20$ layers with $26.0\% (+3.5)$ CRA ($\epsilon=36/255$), $15.5\% (+8.0)$ CRA ($\epsilon=108/255$) and $32.6\% (+0.5)$ clean accuracy on \textit{LipConv-20}.
These CRAs outperform the recent best methods \textit{SLL} and \textit{LBDN}.
Differently to the AOL loss used in \textit{SLL} and \textit{LBDN} \ac{ours} also enables soft-constrained architectures like \textit{8C2F} to achieve sota-performances.
\Ie{}, when choosing a lower slack value $p=0.025$ and $\epsilon_{\textit{train}}=1.0$, \ac{ours} on \textit{8C2F} out-competes even \textit{LBDN}~\cite{lbdn}.
\Ie{}, we increase CRAs for $\epsilon=36/255$ to $25.3\% (+0.6)$ and for $\epsilon=108/255$ to $13.8\% (+0.4)$ while maintaining clean accuracy $33.5\% (+0.1)$.
Interestingly, \textit{8C2F} has fewer parameters than \textit{Sandwich} and \textit{XL}: $4.3M$ vs $39M$ vs $1.1B$~\cite{lbdn},

Next, we consider CIFAR-10 and CIFAR-100 (table~\ref{tbl:results})
GloRo applied to \textit{6C2F} and \textit{4C3F} produces a CRA ($\epsilon=36/255$) of under $60\%$ ($58.4\%$ and $59.6\%$ respectively) on CIFAR-10. 
Note that our reimplementation of GloRo improves CRA to $59.6\% (+1.2)$ upon the reported baseline in~\cite{gloro}.
Replaced with \ac{ours}, we report gains to $61.3\% (+2.9)$ and $61.4\% (+1.8)$ respectively.
Additionally, we increase clean accuracy to $77.6\%$ for both ($+0.6$ and $+0.2$ respectively).
Thereby, outperforming the local Lipschitz bound extension \textit{Local-Lip-B}\cite{huang2021locallipb} ($60.7\%$ CRA), which utilizes expensive sample dependent local Lipschitz bounds to increase tightness.
We also compare to \textit{SOC}~\cite{soc}, \textit{CPL}~\cite{cpl} and \textit{SLL}~\cite{sll}, which constrain all layers to have Lipschitz constant $1$. 
Here, we consider the \textit{LipConv-20} architecture for \textit{SOC} and the \textit{XL}-architectures for \textit{CPL} and \textit{SLL}.
\textit{LipConv-20} contains $20$ layers and \textit{XL} $85$.
\textit{SOC} achieves a clean accuracy of $76.3\%$ on CIFAR-10, which \ac{ours} improves to $77.4\% (+1.1)$ while also improving CRA on all tested margins.
\Eg{} for $\epsilon=36/255$ we report a gain to $64.2\% (+1.6)$ and for $\epsilon=128/255$ a gain to $36.7\% (+0.7)$.
Similarly, when applying \ac{ours} to \textit{CPL-XL}, we report CRA gains to $65.9\% (+1.5)$ and $38.1\% (+5.1)$ ($\epsilon=36/255$ and $\epsilon=128/255$ respectively) while retaining clean accuracy ($+0.3$).
However, \ac{ours} on \textit{SLL-XL} provides no improvements.
This is due to \textit{SLL} using the \textit{AOL}-loss~\cite{aol}, which has similar properties to \ac{ours} on $K$=$1$ constrained models.
We provide a discussion in the supplement, sec.~D.
On CIFAR-100 though, \ac{ours} provides improvements on all tested models.
On \textit{CPL-XL}, we improve CRA on $\epsilon=108/255$ to $21.5\% (+8.9)$ while retaining clean accuracy $47.9\% (+0.1)$.
On \textit{SLL}, we report slight gains in clean accuracy to $46.9\% (+0.4)$ and CRA ($+0.1$ for both $\epsilon=36/255$ and $\epsilon=128/255$).

\begin{figure*}[t]
    \centering
    \begin{subfigure}[t]{0.24\textwidth}
        \includegraphics[width=\textwidth, trim={0.5cm 0.44cm 0.5cm 0.5cm},clip]{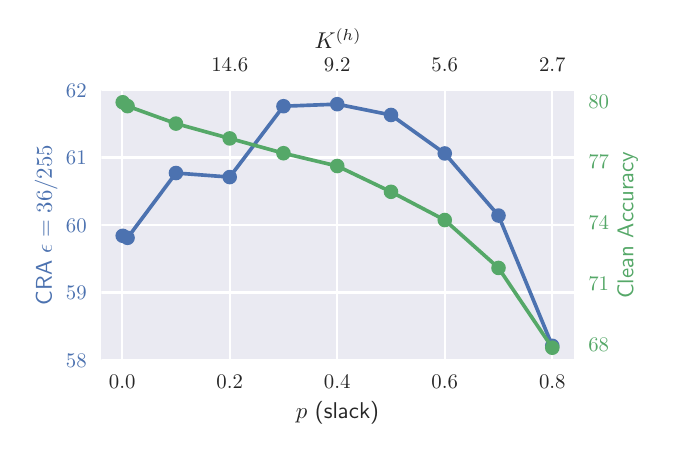}
    \end{subfigure}
    \hfill
    \begin{subfigure}[t]{0.24\textwidth}
        \raisebox{1em}
        \centering
        \resizebox{0.99\textwidth}{!}{
        \begin{tabular}[b]{|@{\hskip 4px}c@{\hskip 4px}|@{\hskip 4px}c@{\hskip 4px}|@{\hskip 4px}c@{\hskip 4px}|}
        \hline
        $\epsilon_\text{train}$ & GloRo & Ours\\
        \hline
        0.15 & 58.8 & 60.6 (\textbf{+3.0$\%$}) \\
        0.25 & 49.6 & 51.6 (\textbf{+4.0$\%$}) \\
        0.50 & 35.9 & 37.0 (\textbf{+3.0$\%$}) \\
        0.75 & 27.0 & 28.7 (\textbf{+6.3$\%$}) \\
        1.00 & 21.2 & 22.5 (\textbf{+6.3$\%$}) \\
        1.25 & 16.9 & 17.4 (\textbf{+3.3$\%$}) \\ \hline
        \end{tabular}
        }
    \end{subfigure}
    \hfill
    \begin{subfigure}[t]{0.24\textwidth}
        \includegraphics[width=\textwidth, trim={0.1cm 0.5cm 0.5cm 0.5cm}, clip]{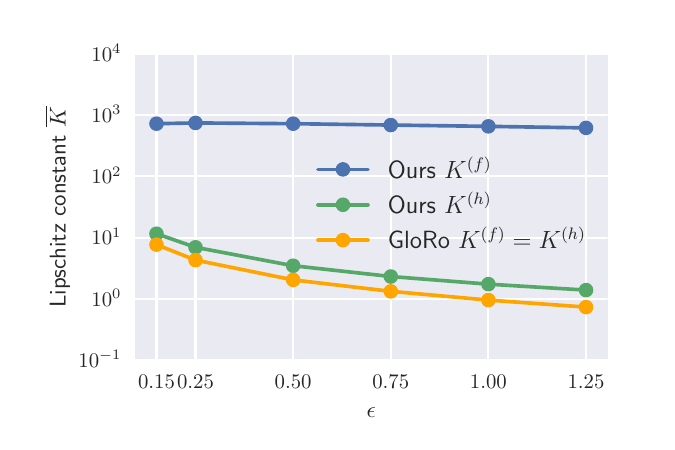}
    \end{subfigure}
    \hfill
    \begin{subfigure}[t]{0.24\textwidth}
        \includegraphics[width=\textwidth, trim={0.1cm 0.5cm 0.5cm 0.5cm}, clip]{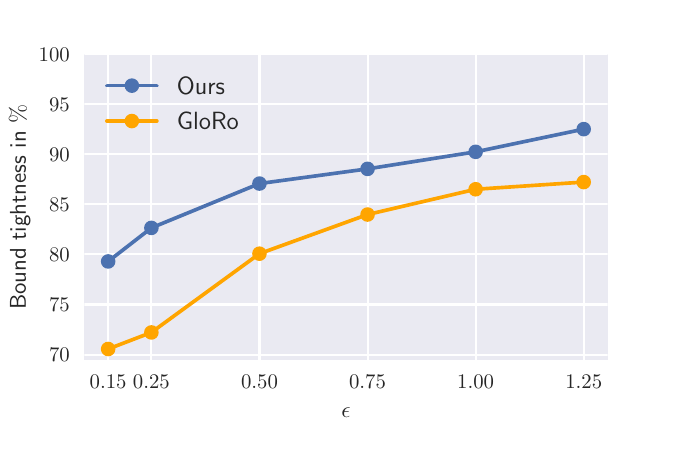}
    \end{subfigure}
    \vspace{-0.2em}
    \caption{
    Left column: Slack governs robust accuracy trade-off in \textit{4C3F} on CIFAR-10.
    Remaining columns: Under increasing $0.15 \leq \epsilon \leq 1.25$ on CIFAR-10, we compare our method with GloRo on \textit{4C3F}. We report consistently better CRA with at least $3\%$ improvement (left table). Our Lipschitz bounds are less constrained and $\kf$ is decoupled from the loss (middle figure). And lastly, our method produces tighter bounds (right).
    \label{fig:eps_sweep}
    }
\end{figure*}

\myparagraph{Lipschitz bound and tightness.}
\ac{ours} offers increased control over the Lipschitz bound of the model $\khf$.
Across all datasets, we find \ac{ours} to increase the Lipschitz constant $\khf$ over the respective baselines (while improving clean and robust accuracies).
On soft-constrained models like \textit{8C2F} on Tiny-ImageNet, \ac{ours} allows a doubling of the constant over GloRo ($7.3$ vs $3.9$).
Similarly, on the hard-constrained model \textit{LipConv-20} on Tiny-Imagenet, we observe another doubling over SOC ($11.6$ vs $6.4$).
This is consistent with $\khf$ on CIFAR.
$\khf$ of \textit{LipConv-20} is increased with \ac{ours} from $5.6$ to $14.7$.
We note an exception for methods that utilize the \textit{AOL}-loss~\cite{aol}: SLL and LBDN.
On these models, we find the constant $\khf$ to be highly similar, \eg{} SLL-XL on CIFAR-100 ($\khf=6$ vs $6.5$).
Importantly, these constant changes come with increased tightness as well.
On Tiny-ImageNet, we increase tightness from $53\%$ to $64\% (+11)$ on \textit{8C2F} over GloRo and from $81\%$ to $84\% (+3)$ on \textit{LipConv-20} over SOC.
Similarly, on CIFAR-10, we increase tightness from $71\%$ to $80\%$ on \textit{4C3F} over GloRo.
In general, we observe the largest tightness improvements on soft-constrained models, although improvements on \textit{CPL} are substantial.
Here we gain, $+2$ and $+5$ percent points on CIFAR-10 and CIFAR-100 respectively.

\subsection{Analysis and ablation}
\label{sec:evaluation:analysis}
We considered different design choices when training with \ac{ours}.
An important aspect being the robust accuracy trade-off, which we investigate in the following by controlling slack.
Furthermore, we investigate the bound, tightness and CRA over increasing margin width on CIFAR-10.
An extended discussion on selecting $\epsilon$ and $p$ is discussed on \textit{8C2F} for Tiny-ImageNet in the supplement, sec~D.

\myparagraph{Slack.}
As discussed in section~\ref{sec:method:slack}, we can regard the calibration probability $p$ as slack, which trades-off CRA and clean accuracy.
We report both on \textit{4C3F} for $p \in [0.01, 0.8]$ in figure~\ref{fig:eps_sweep} (left).
An increase in $p$ decreases clean accuracy (green) from $80\%$ to $68\%$ but CRA increases to a peak at $p=0.4$ with $61.9\%$.

\myparagraph{Increasing margin width.}
In addition to our main results, we compare \ac{ours} and GloRo for larger $\epsilon$.
We train all experiments on \textit{4C3F}.
Different from before, we evaluate not for $\epsilon=\nicefrac{36}{255}$ but for the trained target, such that $\epsilon_\text{train}=\epsilon_\text{test}$.
The table in figure~\ref{fig:eps_sweep}, reports CRA and relative improvement over GloRo. 
With a minimum of $3\%$, we report consistent relative improvement for all $\epsilon$.
The two right most plots of figure~\ref{fig:eps_sweep}, display the Lipschitz constants across $\epsilon$ and tightness respectively.
We see \ac{ours} utilizing higher $\khf$ throughout while maintaining higher tightness.
Note that $\kf$ remains fairly unconstrained at $\approx 10^3$ with \ac{ours}, which is regularized via parameter $\lambda$.
We analyzed its effect by choosing values from $10^{-15}$ to $10^{-1}$ and present results in figure~D4a in the supplement.
We find the performance of the model to be insensitive within $\lambda \in [10^{-10}, 10^{-3}]$.

\section{Conclusion}
We proposed a new loss, \ac{ours}, for Lipschitz margin training that is calibrated to the margin width.
\ac{ours} reveals two intriguing properties: (i) the calibrated distribution width can be interpreted as slackness and (ii) slackness governs the smoothness of the model -- and thereby the Lipschitz constant.
The ramifications are important for improving certified robustness with Lipschitz constants.
The constant can be large -- implying increased model complexity and accuracy -- if the model is capable of separating the data well.
We illustrated these mechanics on two-moons, highlighting the implications for Lipschitz margin training and provided additional results on CIFAR-10, CIFAR-100 and Tiny-ImageNet.
Applied across a wide range of datasets and methods, \ac{ours} consistently improved clean and certified robustness.

\section*{Acknowledgments}
This work was partially funded by ELSA – European Lighthouse on Secure and Safe AI funded by the European Union under grant agreement No. 101070617. 
Views and opinions expressed are however those of the authors only and do not necessarily reflect those of the European Union or European Commission. Neither the European Union nor the European Commission can be held responsible for them.

\bibliographystyle{splncs04}
\bibliography{main}

\end{document}



\def\SubNumber{087}

\def\GCPRTrack{Fast Review Track}

\title{Certified Robust Models with Slack Control and Large Lipschitz Constants\\\vspace{1em}{\large Supplement}}

\ifreview
\else
	\titlerunning{Supplement - Cert. Rob. Models with Slack Ctrl. and Large Lip. Constants}

	\author{Max Losch\inst{1} \and
	David Stutz\inst{1} \and
	Bernt Schiele\inst{1} \and
    Mario Fritz\inst{2}
    }
	
	\authorrunning{M. Losch et al.}
	
	\institute{$^1$Max Planck Institute for Informatics, Saarland Informatics Campus\\ Saarbr\"ucken, Germany\\
	\email{\{mlosch, dstutz, schiele\}@mpi-inf.com}\\
	$^2$CISPA Helmholtz Center for Information Security\\
    Saarbr\"ucken, Germany\\
    \email{fritz@cispa.de}}
\fi

\appendix
\maketitle              

In this supplement, we provide additional material on our \acf{ours}.
The order of presentation is aligned with the order in the main paper. 
That is, we start with material on constructing~\ac{ours} in section~\ref{appendix:ours} and continue with discussing implementation and training setups in section~\ref{appendix:implementation} followed by used evaluation metrics in section~\ref{appendix:metrics}.
We conclude with additional experiments in section~\ref{appendix:experiments}.
Here, we provide additional analysis on Two-Moons and CIFAR-10 among which we discuss the regularization coefficient $\lambda$ in more detail and show that a GloRo loss with fixed temperature scaling does not endow cross-entropy with control over slack. 

\newcommand{\additem}[2]{%
\item[\textbf{(\ref{#1})}] 
    \textbf{#2} \dotfill\makebox{\textbf{\pageref{#1}}}
}

\newcommand{\addsubitem}[2]{%
    \\[.5em]\indent\hspace{1em}
    \textbf{(\ref{#1})}
    #2 \dotfill\makebox{\textbf{\pageref{#1}}}
}

\newcommand{\adddescription}[1]{\newline
\begin{adjustwidth}{1cm}{1cm}
#1
\end{adjustwidth}
}

\begin{enumerate}[label={({\arabic*})}, topsep=1em, itemsep=.5em]
    \additem{appendix:ours}{ 
    \acf{ours}} 
%
%
    \additem{appendix:implementation}{ 
    Implementation and Setup} 
%
%
    \additem{appendix:metrics}{ 
    Metrics} 
%
%
    \additem{appendix:experiments}{ 
    Additional experiments} 
%
%
\end{enumerate}

\section{\acf{ours}}
\label{appendix:ours}

The main paper discussed the calibration of the logistic function for the binary case.
To complete this discussion, we provide its generalization to the multiclass case, provide a derivation of $\sigma$  (equation~5 in the main paper) and a proof for $N=2$ that our multiclass~\ac{ours} remains a generalization of the logistic function.
Additionally, we delineate \ac{ours} from the \textit{AOL}-loss~\cite{aol} which provides similar properties on $K=1$-constrained models.

\myparagraph{Multiclass \ac{ours}.}
\newcommand{\ind}{\mathbbm{1}}
The general intuition developed in the binary case above can readily be applied to multiclass classification. 
The main difference being the application to pairs of output logits (before we considered a scalar output).
To this end, let $y\in \{1, 2, ..., N\}$ and $f: \mathcal{X} \to \mathbb{R}^N$.
To calibrate a classifier with multinomial output, we need to measure the output distances between pairs of logits, e.g. $f_{i,y}(x) = f_i(x)-f_y(x)$ for logit $i$ and logit of label $y$.
We denote the Lipschitz constant of $f_{i,y}(x)$ as $K_{i,y}$ and generalize binary-\ac{ours} to the cross-entropy loss.
To construct \ac{ours}, we make use of softmaxs shift invariance property -- any constant can be subtracted from the input without changing the output.
That is, by subtracting $f_y(x)$ from all output logits, we do not change the softmax output distribution and we can calibrate the margin widths for all one-vs-one decision boundaries.

\myparagraph{Definition 2 - Multinomial \ac{ours}.}
\textit{Let $\mathcal{L}$ be the negative log-likelihood loss $\mathcal{L}(y, h(f(x);y))=-\log h(f(x);y)$, then the calibrated softmax is defined as follows:}
\begin{align}
    \hat{h}(f(x);y) = \frac{1}{\sum_{i=1}^N e^{g_y(i,x)}} 
    \begin{bmatrix}
    e^{g_y(1,x)}\\
    e^{g_y(2,x)}\\
    \smash{\vdots} \\
    e^{g_y(N,x)}\\
    \end{bmatrix}\label{eq:ce_full}\\
    \,\text{ \text{with} }\,
    g_y(i,x) = \begin{cases}
    \frac{\epsilon}{\sigma_{\epsilon}(p)}+\frac{f_{i,y}(x)}{\sigma_{\epsilon}(p) K_{i,y}} &\text{if $i\neq y$}\\
    -\frac{\epsilon}{\sigma_{\epsilon}(p)} &\text{if $i = y$}\\
    \end{cases}.
\end{align}
Here, the core implementation is $g_y(i,x)$.
Equal to binary-\ac{ours}, the multinomial formulation retains the same properties.
That is, (i) normalization by $K$, and (ii): calibration with $\sigma_\epsilon(p)$.
Since we consider logit differences $f_{i,y}$ we obtain two cases: $i\neq y$ and $i=y$.
For the latter, $f_{y,y}$ equals $0$ for all $x$, hence $g_y(y,x)$ reduces to 
$-\nicefrac{\epsilon}{\sigma_\epsilon(p)}$.
To assert, equation~\ref{eq:ce_full} is still a generalization of the logistic function, we provide proof when $N=2$ next.

It is well known that softmax is a generalization of the logistic distribution.
The generalization can be shown easily when the number of logits $N$ is $2$.
Here, we show, that our calibrated softmax for \ac{ours} remains a logistic generalization.
We make use of the shift invariance of softmax -- any constant can be added or subtracted without changing the output.
For simplification, let $\sigma_\epsilon(p)$ simply be $\sigma$ and $k=1/K^{\spow{f}}_{1,2}=1/K^{\spow{f}}_{2,1}$ and $a=\nicefrac{\epsilon}{\sigma}$.
The multiclass-\ac{ours} for $N=2$ is then defined as
\begin{align}
    \hat{h}(f(x);y) = \frac{1}{e^{g_y(1,x)} + e^{g_y(2,x)}} 
    \begin{bmatrix}
    e^{g_y(1,x)}\\
    e^{g_y(2,x)}\\
    \end{bmatrix}
    \\\text{ \textit{with} }\,
    g_y(i,x) = \begin{cases}
    a+\nicefrac{k}{\sigma} f_{i,y}(x) &\text{if $i\neq y$}\\
    -a &\text{if $i = y$}\\
    \end{cases}.
\end{align}

Due to softmaxs shift invariance, we add $a$ to both sides and acquire:
\begin{align}
    g_y(i,x) = \begin{cases}
    2a+\nicefrac{k}{\sigma}f_{i,y}(x) &\text{if $i\neq y$}\\
    0 &\text{if $i = y$}\\
    \end{cases}.
\end{align}
This formulation encourages a margin width of $2\epsilon$ by adding to all logits where $i\neq y$.
We note that a similar (uncalibrated) variant to this formulation has already been used in Lipschitz Margin Training (LMT)~\cite{LMT}.
In the following, we develop for both cases of $y \in \{1,2\}$.
Let $y=2$, then we have $g_{y=2}(2,x)=0$. 
Otherwise when $y=1$, we have $g_{y=1}(1,x)=0$.
It follows:
\begin{align}
    \hat{h}(f(x);y=2) &= \frac{1}{e^{2a+\nicefrac{k}{\sigma}f_{1,2}(x)} + 1} 
    \begin{bmatrix}
    e^{2a+\nicefrac{k}{\sigma}f_{1,2}(x)}\\
    1
    \end{bmatrix}\\
    &= \begin{bmatrix}
    1-\frac{1}{1+e^{2a+\nicefrac{k}{\sigma}f_{1,2}(x)}}\\
    \frac{1}{1+e^{2a+\nicefrac{k}{\sigma}f_{1,2}(x)}}
    \end{bmatrix}\\
    \hat{h}(f(x);y=1) &= \frac{1}{1 + e^{2a+\nicefrac{k}{\sigma}f_{2,1}(x)}} 
    \begin{bmatrix}
    1\\
    e^{2a+\nicefrac{k}{\sigma}f_{2,1}(x)}
    \end{bmatrix}\\
    &= \begin{bmatrix}
    \frac{1}{1+e^{2a+\nicefrac{k}{\sigma}f_{2,1}(x)}}\\
    1-\frac{1}{1+e^{2a+\nicefrac{k}{\sigma}f_{2,1}(x)}}
    \end{bmatrix}.
\end{align}
We observe, that the representations for $y=1$ and $y=2$ are the same except for the flipped logit pair: $f_{1,2}$, $f_{2,1}$.
It is easy to see, that both cases are instances of the logistic distribution when we flip the sign of the exponents:
\begin{align}
    \hat{h}(f(x);y=2) &= \begin{bmatrix}
    1-\frac{1}{1+e^{-(-2a-\nicefrac{k}{\sigma}f_{1,2}(x)})}\\
    \frac{1}{1+e^{-(-2a-\nicefrac{k}{\sigma}f_{1,2}(x))}}
    \end{bmatrix}.
\end{align}
This derivation shows that our multiclass-\ac{ours} is a generalization of the logistic function.
Consequently, we can utilize the same method of calibration as for the binary case.

\paragraph{Derivation of $\sigma$.}
We consider the logistic distribution with mean $\mu=0$:
\begin{align}
    f(x;\sigma) &= \frac{e^{-\frac{x}{\sigma}}}{\sigma(1+e^{-\frac{x}{\sigma}})^2},\\
    \text{with cdf }\,\, \Phi(x;\sigma)&= h(x;\sigma) = \frac{1}{1+e^{-\frac{x}{\sigma}}}
\end{align}
and want to determine the parameter $\sigma$.
To this end, let $X$ be the random variable of this distribution and let $x_1, x_2$ be two values, $x_1 < x_2$.
Then the probabilities for these values are given by $P(X < x_1) = p_1$ and $P(X < x_2) = p_2$.
The logistic distribution belongs to the location-scale family, which enables reparametrization of $X$:
let $X$ have zero-mean (as assumed) and unit variance.
Then we can define a new random variable $Y:=\sigma X$, which has the cdf $\Phi_Y(y)=\Phi \left(\frac{y}{\sigma}\right)$.
From this follows a set of linear equations:
\begin{align}
    \Phi^{-1}(p_1)\sigma &= x_1\\
    \Phi^{-1}(p_2)\sigma &= x_2
\end{align}
and the corresponding solution used for calibrating~\ac{ours}:
\begin{align}
    \sigma = \frac{x_2-x_1}{\Phi^{-1}(p_2) - \Phi^{-1}(p_1)}.
\end{align}

\subsection{Similarity to \textit{AOL} loss.}
As mentioned in section~4 (main paper), applying \ac{ours} to the \textit{SLL} framework does not improve clean or robust accuracies by the same margins as on the other methods.
This is because, the used \textit{AOL} loss~\cite{aol} has similar properties to \ac{ours} when used on $K=1$ constrained models.
AOL is defined as: $h\left((-uy + f(x)) / t\right) t$,
where $uy$ is an offset similar to $y\epsilon$ and $t$ a normalization factor similar to $\sigma_\epsilon(p)K$.
Importantly, $t$ is fixed throughout training, while $K$ in \ac{ours} is not.
This crucial difference is key: $K$ may change in the orders of magnitudes during training and thus must be accounted for appropriately to ensure calibration.
\ac{ours} is thus a generalization of the \textit{AOL} loss.
This has an especially large effect on soft constrained models, as we saw on experiments on two-moons (figures~4 (main paper) and~\ref{fig:appendix:moons}) and Tiny-ImageNet (table~1, main paper).

\section{Implementation and Setup}
\label{appendix:implementation}
Implementation of our method involves computing the upper Lipschitz bound and training with our calibrated loss.
The latter is discussed on different training datasets as listed in the main paper.
We list implementation details for two-moons as well as CIFAR-10 and Tiny-ImageNet.

\myparagraph{Computing upper Lipschitz bound.}
We implement the upper bound defined in equation~\eqref{eq:lip_upper_bound} by utilizing the product of spectral norms~\cite{szegedy2014}.
We follow a common approach in calculating the spectral norm by performing power iteration and closely follow the strategy in GloRo~\cite{gloro}.
That is, initialize the power iterates independently for each layer, update $5$ times for each forward pass and reuse the state.
\cite{huang2021locallipb} found this to work best over random re-initializations.
To ensure the bound is exact during evaluation, we let the power iteration converge once after each training epoch until the error between iterations is smaller than $10^{-9}$.
During validation and testing, the calculated spectral norms are kept fixed.
\begin{equation}
    K \leq \prod_{l=1}^{L} \|g^{(l)}\|_p =: \hat{K}.
    \label{eq:lip_upper_bound}
\end{equation}

\myparagraph{Two-moons setup.}
For our two-moons experiments, we generate the data via the scikit-learn~\cite{scikit-learn} package, but use uniform sampled noise instead of normal distributed noise.
For each point we sample within a circle of radius $0.1$, such that the resulting true-margin is exactly $2\epsilon = 0.3$.
The training data is generated with $2000$ samples.

On this dataset, we train $7$-layered MLPs with MinMax activations.
All models have two output logits and are trained with variants of the cross-entropy loss.
GloRo instruments an additional logit and our method utilizes our calibrated version in equation~\eqref{eq:ce_full}.
We set $p$ to $10^{-12}$ in \ac{ours} to decrease slack and increase $K$.
We found training to converge faster when annealing $p$ from $p_0=10^{-3}$ to $p_T=10^{-12}$.
We utilize polynomial decay 
\begin{align}
    p(t) = p_T + (p_0-p_T) \cdot (1-\nicefrac{t}{T})^\gamma,
    \label{eq:polynomial}
\end{align}
where $t$ denotes the training iteration, $T$ the total number of iterations and $\gamma$ defines the speed of decay.
Our $\gamma$ is set to $50$.
An overview over given hyperparameters for ours and GloRo is given in table~\ref{tab:appendix:moons_config}.
\begin{table*}[h]
\centering
\caption{Two-moons training configuration for GloRo and \ac{ours} (ours).}
\label{tab:appendix:moons_config}
\resizebox{0.8\linewidth}{!}{
\begin{tabular}{l|l|l|l|l|l|l|l|}
\cline{2-8}
 & architecture & epochs & batchsize & lr & lr decay & $p$ & $\lambda$ \\ \hline
\multicolumn{1}{|l|}{GloRo} & 10-20-40-40-20-10-2 & 1000 & 100 & $10^{-3}$ & to $10^{-4}$ at epoch 400 & N/A & N/A \\ \hline
\multicolumn{1}{|l|}{Ours} & 10-20-40-40-20-10-2 & 1000 & 100 & $10^{-3}$ & to $10^{-4}$ at epoch 400 & $10^{-3} \to 10^{-12}$ & $10^{-6}$ \\ \hline
\end{tabular}
}
\end{table*}

\myparagraph{Image dataset setup.}
We trained models on three different image datasets: CIFAR-10, CIFAR-100 and Tiny-ImageNet.
On CIFAR-10, we trained four different architectures: \textit{4C3F}, \textit{6C2F}, \textit{LipConv} and \textit{XL}.
On Tiny-ImageNet, we trained on \textit{8C2F}, \textit{LipConv} and \textit{XL}.
These architectures follow their definitions in their original papers, see~\cite{wong2018scaling},\cite{lee2020lipschitz}, \cite{soc}, \cite{lee2020lipschitz}, \cite{cpl, sll}.
The only difference is in the last layer for $K$=$1$ constrained models used with \ac{ours}.
We use unconstrained final classification layer, instead of last-layer normalization~\cite{soc}.
We found this to provide a slight edge.
We list all architectures trained with \ac{ours} in table~\ref{tab:appendix:arch_configs}.
The syntax in table~\ref{tab:appendix:arch_configs} is as follows:
\textit{c(C,K,S,P)} denotes a convolutional layer with \textit{C} channels, a kernel size of $K\times K$ with a stride \textit{S} and padding \textit{P}.
\textit{d(C)} denotes a fully connected layer with \textit{C} output dimensions.
Note that \textit{8C2F} adds no padding when $S=2$, which is differently to the other architectures.

\begin{table*}[t]
\caption{Architecture configurations on CIFAR-10 (\textit{4C3F}, \textit{6C2F}) and Tiny-ImageNet (\textit{8C2F}).}
\label{tab:appendix:arch_configs}
\resizebox{\linewidth}{!}{
\begin{tabular}{l|l|l|l|l|l|l|l|l|l|l|}
\cline{2-11}
 & \multicolumn{1}{c|}{Layer 1} & \multicolumn{1}{c|}{2} & \multicolumn{1}{c|}{3} & \multicolumn{1}{c|}{4} & \multicolumn{1}{c|}{5} & \multicolumn{1}{c|}{6} & \multicolumn{1}{c|}{7} & \multicolumn{1}{c|}{8} & \multicolumn{1}{c|}{9} & \multicolumn{1}{c|}{10} \\ \hline
\multicolumn{1}{|l|}{4C3F} & c(32,3,1,1) & c(32,4,2,1) & c(64,3,1,1) & c(64,4,2,1) & d(512) & d(512) & d(10) &  &  &  \\ \hline
\multicolumn{1}{|l|}{6C2F} & c(32,3,1,1) & c(32,3,1,1) & c(32,4,2,1) & c(64,3,1,1) & c(64,3,1,1) & c(64,4,2,1) & d(512) & d(10) &  &  \\ \hline
\multicolumn{1}{|l|}{8C2F} & c(64,3,1,1) & c(64,3,1,1) & c(64,4,2,0) & c(128,3,1,1) & c(128,3,1,1) & c(128,4,2,0) & c(256,3,1,1) & c(256,4,2,0) & d(256) & d(200) \\ \hline
\end{tabular}
}
\end{table*}
On CIFAR-10, \textit{LipConv} was trained on the code basis published by Singla \etal{}\cite{soc}  extended with our~\ac{ours} loss.
\textit{4C3F}, \textit{6C2F} and \textit{8C2F} were trained on our own code basis.
The latter three models, were trained for $800$ epochs using a batch size of $128$, except \textit{8C2F} for which we used $256$.
Optimization is done via \textit{Adam}~\cite{kingma2015adam} using no weight decay and a learning rate of $10^{-3}$ on CIFAR-10 and $2.5 \cdot 10^{-4}$ on Tiny-ImageNet, which is decayed by factor $0.1$ at epoch $400$, $600$ and $780$.
In contrast to GloRo, we do not need to anneal the target $\epsilon$ to reach best performances.
On Tiny-ImageNet, we do anneal the normalization with an additional $\beta$ factor $\nicefrac{f(x)}{(K \cdot \beta)}$, where $\beta$ is scheduled with the polynomial formulation in equation~\eqref{eq:polynomial}.
All models utilize the MinMax non-linearity~\cite{anil2019sorting} inplace of ReLU.
All hyperparameters for \textit{4C3F}, \textit{6C2F}, \textit{8C2F} and \textit{LipConv} are summarized in table~\ref{tab:appendix:hyperparams}.

Since \textit{LipConv} had not been evaluated on Tiny-ImageNet in~\cite{soc}, we list performances for a range of $\gamma$ values (the margin regularization) in table~\ref{tab:appendix:soc_on_tinyimagenet}.
We identify $\gamma=0.3$ to work best for \textit{LipConv-10}.
\begin{table}[ht]
    \centering
    \caption{Clean and robust accuracies of \textit{LipConv-10} for different margin influencing parameter $\gamma$. The best configuration ($\gamma=0.3$) is used to compare with \ac{ours} in main table~2 (main paper)}
    \label{tab:appendix:soc_on_tinyimagenet}
    \begin{tabular}{|l|c|c|}
        \hline
        $\gamma$ & Clean (\%) & CRA (\%) \\ \hline
        0.05 & 32.2 & 21.1 \\
        0.15 & 32.1 & 21.4 \\
        0.3 & 32.1 & 21.5 \\
        0.4 & 32.0 & 21.1 \\ 
        \hline
    \end{tabular}
\end{table}

\myparagraph{Data preprocessing.}
For \textit{4C3F}, \textit{6C2F} and \textit{8C2F} we use the following data preprocessing.
All other models use the preprocessing described in their respective papers.
We scale the features to the range $[0,1]$. 
On CIFAR-10, we use the following \textbf{data augmentations} for \textit{4C3F} and \textit{6C2F}:
\begin{enumerate}
\small
    \item \begin{verbatim}random-crop to 32, with padding of size 4\end{verbatim}
    \item \begin{verbatim}random horizontal flip\end{verbatim}
    \item \begin{verbatim}color jitter on each channel: 0.25\end{verbatim}
    \item \begin{verbatim}random-rotation up to 2 degrees\end{verbatim}
\end{enumerate}
On Tiny-ImageNet for \textit{8C2F}, we use exactly the same setup, but increase the crop-size to $64$ with padding of size $8$.

\begin{table*}[t]
\centering
\caption{Hyperparameter configuration for all trained models using \ac{ours}. logarithmic scheduling as described in~\cite{gloro}. TRADES factor is annealed linearly from $0.5$ to $20$, with $20$ reached at epoch $400$.}
\label{tab:appendix:hyperparams}
\resizebox{\textwidth}{!}{
\begin{tabular}{clllllll}
\multicolumn{1}{l}{architecture} & dataset & batch size & epochs & $\epsilon_\text{train}$ & initialization & lr & \\ \hline
\multirow{5}{*}{\begin{tabular}[c]{@{}c@{}}\ac{ours} on\\ \\ 4C3F\\ and\\ 6C2F\end{tabular}} & CIFAR-10 & $128$ & $800$ & $0.15$ & kaiming & $10^{-3}$ & \\
 &  &  &  &  &  &  &  \\ \cline{2-8} 
 & lr-decay & loss & $\epsilon$ schedule & power-iter & $p$ (slack) & $\lambda$ &  \\ \cline{2-8} 
 & $\begin{matrix}10^{-4} & \text{at } 400\\10^{-5} & \text{at } 600\\10^{-6} & \text{at } 780\end{matrix}$ & \ac{ours} & fixed & $5$ & $0.15$ & $10^{-4}$ &  \\
 &  &  &  &  &  &  &  \\ \hline\hline
 \multicolumn{1}{l}{architecture} & dataset & batch size & epochs & $\epsilon_\text{train}$ & initialization & lr & \\ \hline
\multirow{5}{*}{\begin{tabular}[c]{@{}c@{}}\ac{ours} on\\ \\ LipConv\end{tabular}} & CIFAR-10 & $128$ & $200$ & $0.25$ & kaiming & $10^{-3}$ & \\
 &  &  &  &  &  &  &  \\ \cline{2-8} 
 & lr-decay & loss & $\epsilon$ schedule & power-iter & $p$ (slack) & $\lambda$ & LLN \\ \cline{2-8} 
 & $\begin{matrix}10^{-4} & \text{at } 100\\10^{-5} & \text{at } 150\end{matrix}$ & \ac{ours} & fixed & N/A & $5\cdot 10^{-2}$ & $10^{-3}$ & no \\
 &  &  &  &  &  &  &  \\ \hline\hline
\multicolumn{1}{l}{architecture} & dataset & batch size & epochs & $\epsilon_\text{train}$ & initialization & lr \\ \hline
\multirow{5}{*}{\begin{tabular}[c]{@{}c@{}}GloRo on\\ \\ 4C3F\\ and\\ 6C2F\end{tabular}} & CIFAR-10 & $128$ & $800$ & $0.15$ & kaiming & $10^{-3}$ & \\
 &  &  &  &  &  &  &  \\ \cline{2-8} 
 & lr-decay & loss & $\epsilon$ schedule & power-iter &  &  &  \\ \cline{2-8} 
 & $\begin{matrix}10^{-4} & \text{at } 400\\10^{-5} & \text{at } 600\\10^{-6} & \text{at } 780\end{matrix}$ & \begin{tabular}[l]{@{}l@{}}GloRo\\\end{tabular} & logarithmic & $5$ &  &  &  \\
 &  & &  &  &  &  &  \\ \hline\hline
\multicolumn{1}{l}{architecture} & dataset & batch size & epochs & $\epsilon_\text{train}$ & initialization & lr & \\ \hline
\multirow{5}{*}{\begin{tabular}[c]{@{}c@{}}\ac{ours} on\\ \\ 8C2F\end{tabular}} & Tiny-ImageNet & $256$ & $800$ & $0.5$ / $1.0$ & kaiming & $2.5 \cdot 10^{-4}$ & \\
 &  &  &  &  &  &  &  \\ \cline{2-8} 
 & lr-decay & loss & $\epsilon$ schedule & power-iter & $p$ (slack) & $\lambda$ & K schedule \\ \cline{2-8} 
 & $\begin{matrix}2.5\cdot 10^{-5} & \text{at } 400\\2.5\cdot 10^{-6} & \text{at } 600\\2.5\cdot 10^{-7} & \text{at } 780\end{matrix}$ & \ac{ours} & fixed & $5$ & $0.01$ / $0.025$ & $10^{-5}$ & \begin{tabular}[c]{@{}l@{}}polynomial\\$0.01 \to 1.0$, $\gamma=10$\end{tabular} \\
 &  &  &  &  &  &  &  \\ \hline\hline
\multicolumn{1}{l}{architecture} & dataset & batch size & epochs & $\epsilon_\text{train}$ & initialization & lr & \\ \hline
\multirow{5}{*}{\begin{tabular}[c]{@{}c@{}}\ac{ours} on\\ \\ (CPL-)XL\end{tabular}} & CIFAR-10 & $256$ & $200$ & $0.25$ & kaiming & $10^{-3}$ & \\
 &  &  &  &  &  &  &  \\ \cline{2-8} 
 & lr-schedule & loss & $\epsilon$ schedule & power-iter & $p$ (slack) & $\lambda$ & LLN \\ \cline{2-8} 
 & triangular\cite{cpl} & \ac{ours} & fixed & N/A & $0.1$ & $10^{-3}$ & no \\
 &  &  &  &  &  &  &  \\ \hline\hline
\multicolumn{1}{l}{architecture} & dataset & batch size & epochs & $\epsilon_\text{train}$ & initialization & lr & \\ \hline
\multirow{5}{*}{\begin{tabular}[c]{@{}c@{}}\ac{ours} on\\ \\ (CPL-)XL\end{tabular}} & CIFAR-100 & $256$ & $200$ & $1.0$ & kaiming & $10^{-3}$ & \\
 &  &  &  &  &  &  &  \\ \cline{2-8} 
 & lr-schedule & loss & $\epsilon$ schedule & power-iter & $p$ (slack) & $\lambda$ & LLN \\ \cline{2-8} 
 & triangular\cite{cpl} & \ac{ours} & fixed & N/A & $10^{-3}$ & $10^{-3}$ & no \\
 &  &  &  &  &  &  &  \\ \hline\hline
\multicolumn{1}{l}{architecture} & dataset & batch size & epochs & $\epsilon_\text{train}$ & initialization & lr & \\ \hline
\multirow{5}{*}{\begin{tabular}[c]{@{}c@{}}\ac{ours} on\\ \\ (SLL-)XL\end{tabular}} & CIFAR-10 & $256$ & $1000$ & $1.0$ & kaiming & $10^{-2}$ & \\
 &  &  &  &  &  &  &  \\ \cline{2-8} 
 & lr-schedule & loss & $\epsilon$ schedule & power-iter & $p$ (slack) & $\lambda$ & LLN \\ \cline{2-8} 
 & triangular\cite{sll} & \ac{ours} & fixed & N/A & $5\cdot 10^{-5}$ & $10^{-3}$ & no \\
 &  &  &  &  &  &  &  \\ \hline\hline
\multicolumn{1}{l}{architecture} & dataset & batch size & epochs & $\epsilon_\text{train}$ & initialization & lr & \\ \hline
\multirow{5}{*}{\begin{tabular}[c]{@{}c@{}}\ac{ours} on\\ \\ (SLL-)XL\end{tabular}} & CIFAR-100 & $256$ & $1000$ & $1.0$ & kaiming & $10^{-2}$ & \\
 &  &  &  &  &  &  &  \\ \cline{2-8} 
 & lr-schedule & loss & $\epsilon$ schedule & power-iter & $p$ (slack) & $\lambda$ & LLN \\ \cline{2-8} 
 & triangular\cite{sll} & \ac{ours} & fixed & N/A & $3 \cdot 10^{-3}$ & $10^{-3}$ & no \\
 &  &  &  &  &  &  &  \\ \hline\hline
\multicolumn{1}{l}{architecture} & dataset & batch size & epochs & $\epsilon_\text{train}$ & initialization & lr & \\ \hline
\multirow{5}{*}{\begin{tabular}[c]{@{}c@{}}\ac{ours} on\\ \\ LipConv\end{tabular}} & Tiny-ImageNet & $128$ & $200$ & $0.15$ &  kaiming & $10^{-3}$ & \\
 &  &  &  &  &  &  &  \\ \cline{2-8} 
 & lr-decay & loss & $\epsilon$ schedule & power-iter & $p$ (slack) & $\lambda$ & LLN \\ \cline{2-8} 
 & $\begin{matrix}10^{-4} & \text{at } 100\\10^{-5} & \text{at } 150\end{matrix}$ & \ac{ours} & fixed & N/A & $3\cdot 10^{-1}$ & $10^{-5}$ & no \\
 &  &  &  &  &  &  &  \\ \hline\hline
\multicolumn{1}{l}{architecture} & dataset & batch size & epochs & $\epsilon_\text{train}$ & initialization & lr & \\ \hline
\multirow{5}{*}{\begin{tabular}[c]{@{}c@{}}GloRo on\\ \\ 8C2F\end{tabular}} & Tiny-ImageNet & $256$ & $800$ & $0.15$ & kaiming & $10^{-4}$ & \\
 & \multicolumn{1}{c}{} &  &  &  &  &  &  \\ \cline{2-8} 
 & lr-decay & loss & $\epsilon$ schedule & power-iter &  &  &  \\ \cline{2-8} 
 & $\begin{matrix}10^{-5} & \text{at } 400\\10^{-6} & \text{at } 600\\10^{-7} & \text{at } 780\end{matrix}$ & \begin{tabular}[c]{@{}c@{}}GloRo + TRADES\\$0.5 \to 20$ at ep400\end{tabular} & logarithmic & $5$ &  &  &  \\
 &  &  &  &  &  &  &  \\ \hline
\end{tabular}
}
\end{table*}

\section{Metrics}
\label{appendix:metrics}

\myparagraph{Certified robust accuracy (CRA).}
To compare models, we measure the number of accurately classified samples that also fall outside the margin--hence are certified $\epsilon$-robust.
Certification of input samples is straight forward with Lipschitz models.
That is, an input sample is certified when the smallest difference between highest predicting logit $f_i$ and any other logit $f_j$ is greater than $\epsilon\kf_{i, j}$:
\begin{align}
    \mathcal{M}_f(x) > \epsilon \kf_{i, j}\,\,,\text{where } \mathcal{M}_f(x) = f_i(x)-\max_{j\neq i} f_j(x)
\end{align}
We report the total of certified samples that are also accurately classified as CRA.
For simplicity, we perform CRA evaluation on the classifier $f$ without $\hat{h}$.

\myparagraph{Tightness.}
Recall that the exact Lipschitz constant is NP-hard to determine (section~3.1, main paper).
As remedy, we resort to upper bounds, for which high tightness is important to avoid overestimation of output distances.
In order to demonstrate improved tightness using our loss over existing alternatives, we measure this tightness by estimating an empirical lower bound $\kf_\text{lobo}$ on the training dataset, such that $\kf_\text{lobo} \leq \kf \leq \hat{K}^\spow{f}$.
We state the tightness $\nicefrac{\kf_\text{lobo}}{\hat{K}^\spow{f}}$ in percent.
Note that while this ratio is an approximation, the true tightness is strictly greater or equal than our reported quantity.
To estimate $\kf_\text{lobo}$, we follow the description in~\cite{gloro}:
\begin{align}
    \max_{x_1,x_2} \max_i \left\{ \frac{|f_{j_1}(x_1) - f_i(x_1) - (f_{j_1}(x_2) - f_i(x_2)|}{||x_1-x_2||_2} \right\},
    \label{eq:lobo}
\end{align}
where $j_1 = \arg\max_i f_i(x_1)$.
Intuitively, we sample two inputs $x_1, x_2$, and perturb both to maximize the quotient which is a lower bound to $K_{i,j}$:
\begin{align}
    \frac{|f_{j}(x_1) - f_i(x_1) - (f_{j}(x_2) - f_i(x_2))|}{||x_1-x_2||_2} \leq K_{i,j}
\end{align}
To optimize equation~\eqref{eq:lobo}, we use the \textit{Adam} optimizer and perform $1000$ iterations with a learning rate of $3\cdot 10^{-4}$.
We sample $4000$ pairs from the training dataset and report the largest value obtained over all pairs.

\section{Additional experiments}
\label{appendix:experiments}

\myparagraph{Slack control on two-moons.}
In our discussion on two-moons we have highlighted the connection between slack and Lipschitz constant.
Specifically, we showed that choosing low slack can train models with perfect CRA.
Here, we show that choosing too much slack does indeed result in increased classifier smoothness.
We follow the setup in the main paper but train only for $2\epsilon=0.3$ -- the true margin width.
The results for decreasing $p$ from $0.75$ to $0.0001$ are shown in figure~\ref{fig:appendix:moons}.
While the smallest shown $p=0.0001$ nearly achieves perfect CRA, a higher slack of $p=0.25$ produces very similar results to GloRo (see figure~4b, main paper), with training points remaining within the margins.
Increasing slack even further $p=0.75$, we see a near piecewise-linear classifier, unsuitable to generalize to the data -- as discussed in section~3.3 (main paper).

\myparagraph{SOC on two-moons.}
In the main paper we evaluated GloRo and \ac{ours} on two-moons to highlight the influence of $K$ on classifier smoothness.
For completeness, we additionally present results with 1-Lipschitz constrained models in figure~\ref{fig:appendix:moons_soc}.
We train models of three different depths: $3$, $7$ and $25$ (in columns) using the \textit{SOC} method~\cite{soc}.
\textit{SOC} employs the use of a margin regularization parameter $\gamma$, which we evaluate for three different values: $0.0$, $0.3$ and $1.0$ (in rows).
Overall, we find none of the tested configurations perform better than $90\%$ CRA.
All decision functions are too smooth to fit the data, while the Lipschitz constant $\khf=2.0$ is very low.
These observations are in line with our main paper discussions on the link between $K$ and classifier complexity.

\subsection{Analysis and ablation details.}

\begin{figure}[ht]
    \centering
    \includegraphics[width=0.4\columnwidth, trim={0.4cm 0 0.2cm 0.2cm},clip]{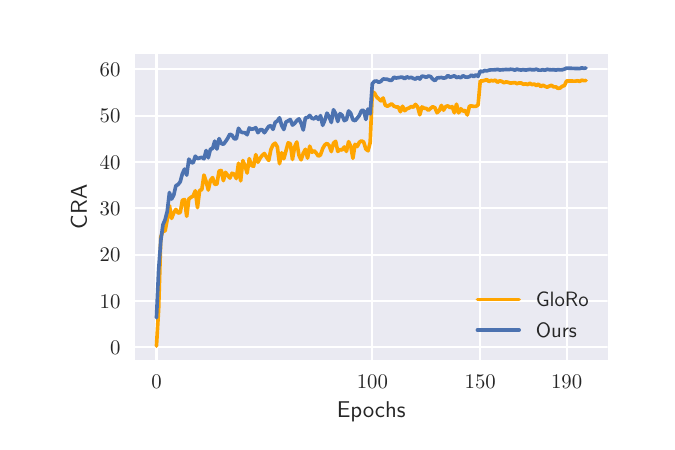}
    \vspace{-1em}
    \caption{Our loss converges faster. Here for \textit{4C3F} on CIFAR-10.}
    \label{fig:convergence}
\end{figure}
For training~\ac{ours} we considered different hyperparameters in order to maximize performance.
In the following, we provide additional material on convergence speed improvements over GloRo, finding optimal values for the regularization coefficient $\lambda$ on CIFAR-10 and evaluating the impact of slack ($p$) on CRA on the constrained \textit{LipConv} architectures.
Finally, we show empirical evidence that existing Lipschitz margin losses like GloRo do not possess slack control.

\begin{figure*}
    \begin{minipage}[t]{0.014\textwidth}
        \vspace{4.5em}
        \rotatebox[origin=c]{90}{Ours}
    \end{minipage}
    \begin{subfigure}[t]{0.235\textwidth}
        \centering
        \caption{$p = 0.75$}
        \label{fig:appendix:moons_a}
        \includegraphics[width=\textwidth]{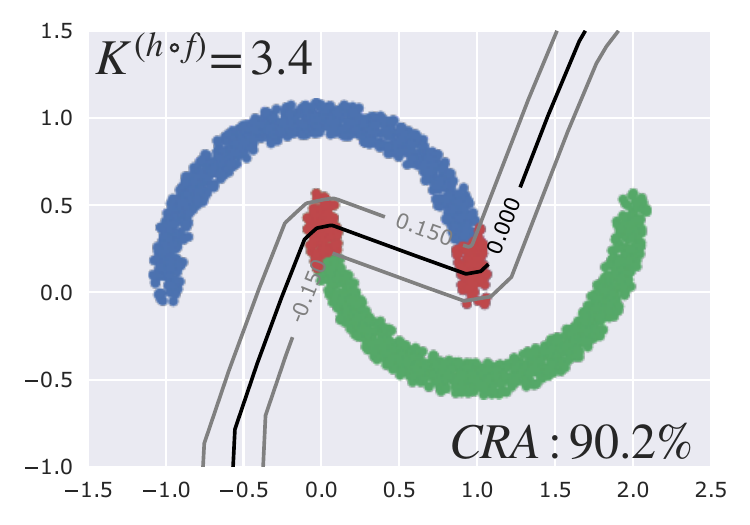}
    \end{subfigure}
    \hfill
    \begin{subfigure}[t]{0.235\textwidth}
        \centering
        \caption{$p = 0.25$}
        \label{fig:appendix:moons_b}
        \includegraphics[width=\textwidth]{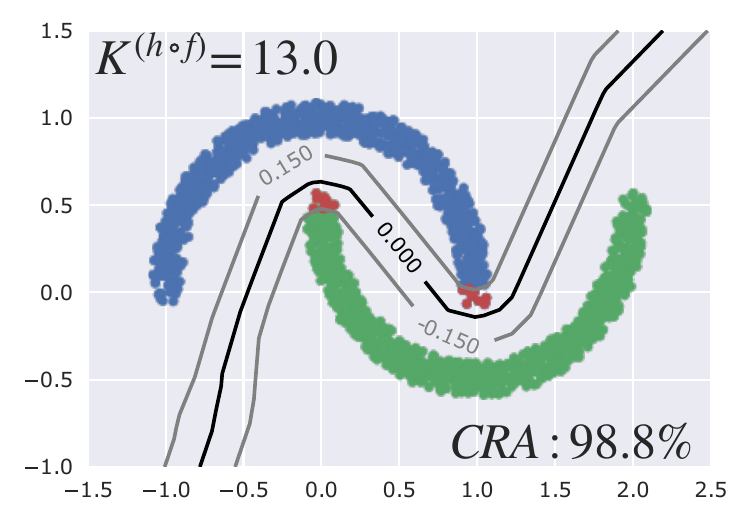}
    \end{subfigure}
    \hfill
    \begin{subfigure}[t]{0.235\textwidth}
        \centering
        \caption{$p = 0.01$}
        \label{fig:appendix:moons_c}
        \includegraphics[width=\textwidth]{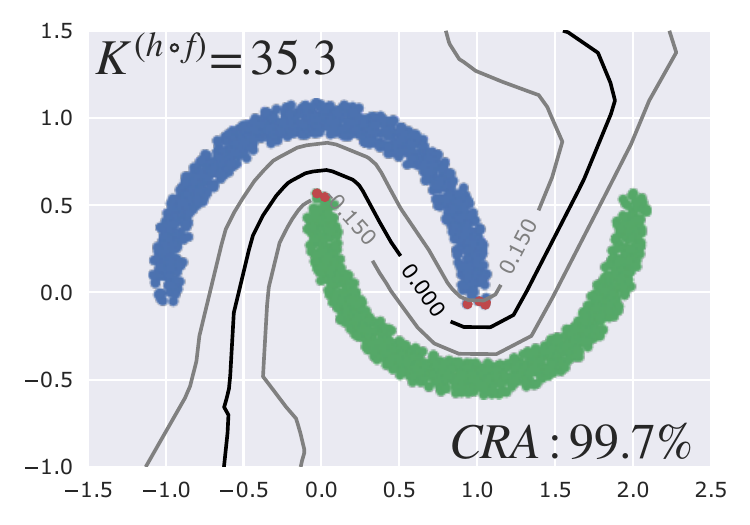}
    \end{subfigure}
    \hfill
    \begin{subfigure}[t]{0.235\textwidth}
        \centering
        \caption{$p = 0.0001$}
        \label{fig:appendix:moons_d}
        \includegraphics[width=\textwidth]{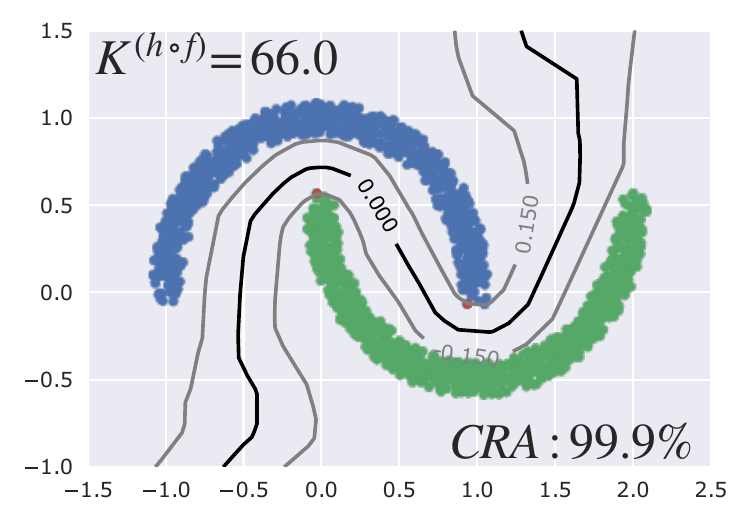}
    \end{subfigure}
    \hfill
    \caption{Slack control on two moons with \ac{ours}. With high $p$ and thus high slack, the decision boundary becomes overly smooth. To reach a perfect CRA score as shown in the main paper, $p$ needs to be smaller than $0.0001$ (right most figure).}
    \label{fig:appendix:moons}
\end{figure*}

\begin{figure*}
    \centering
    \begin{minipage}[t]{0.014\textwidth}
        \vspace{3.5em}
        \rotatebox[origin=c]{90}{$\gamma = 0.0$}
    \end{minipage}
    \begin{subfigure}[t]{0.24\textwidth}
        \centering
        \caption{3 layers}
        \includegraphics[width=\textwidth]{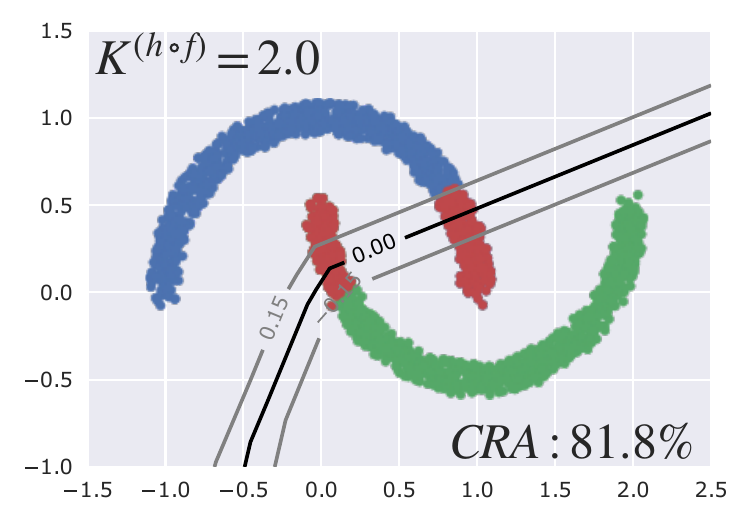}
    \end{subfigure}
    \begin{subfigure}[t]{0.24\textwidth}
        \centering
        \caption{7 layers}
        \includegraphics[width=\textwidth]{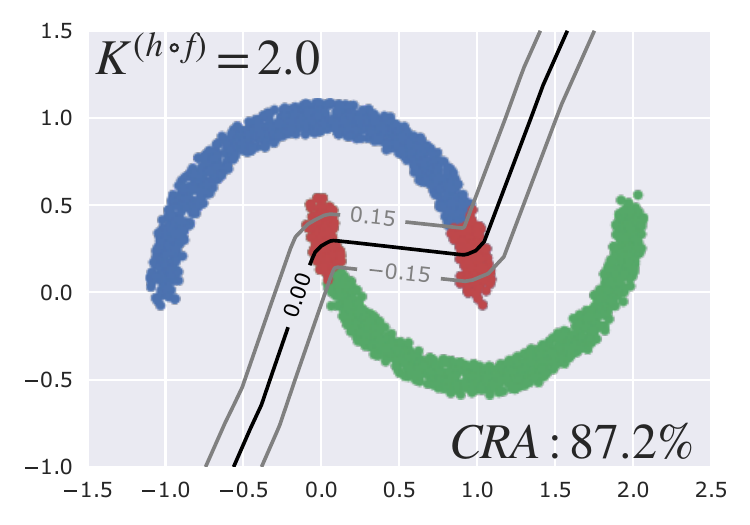}
    \end{subfigure}
    \begin{subfigure}[t]{0.24\textwidth}
        \centering
        \caption{25 layers}
        \includegraphics[width=\textwidth]{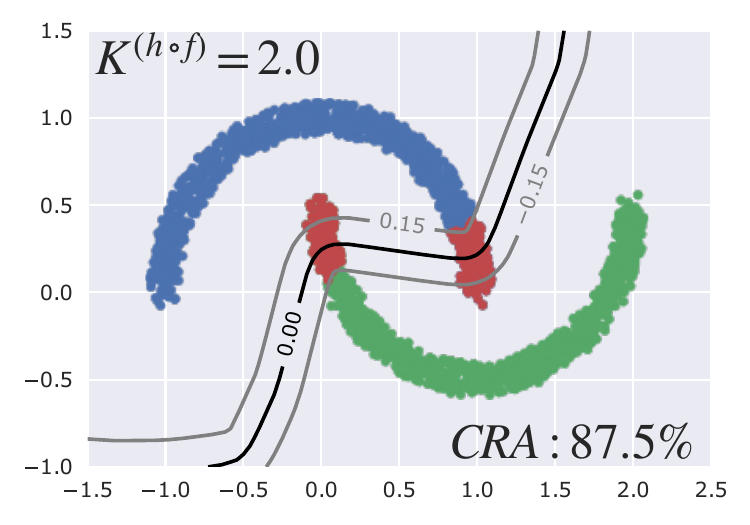}
    \end{subfigure}
    \hfill
    
    \begin{minipage}[t]{0.014\textwidth}
        \vspace{2.5em}
        \rotatebox[origin=c]{90}{$\gamma = 0.3$}
    \end{minipage}
    \begin{subfigure}[t]{0.24\textwidth}
        \centering
        \vspace{-0.1em}
        \includegraphics[width=\textwidth]{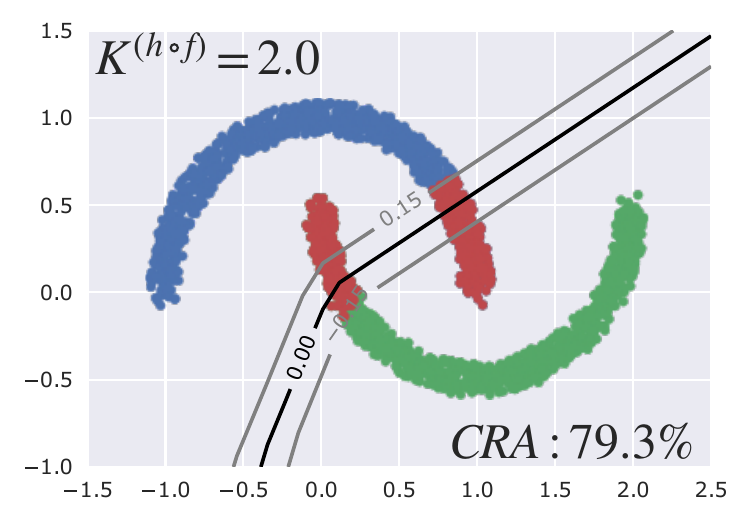}
    \end{subfigure}
    \begin{subfigure}[t]{0.24\textwidth}
        \centering
        \vspace{-0.1em}
        \includegraphics[width=\textwidth]{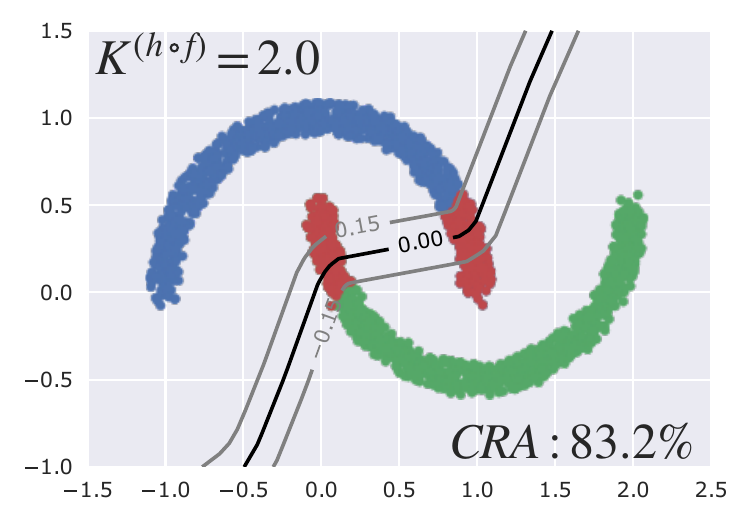}
    \end{subfigure}
    \begin{subfigure}[t]{0.24\textwidth}
        \centering
        \vspace{-0.1em}
        \includegraphics[width=\textwidth]{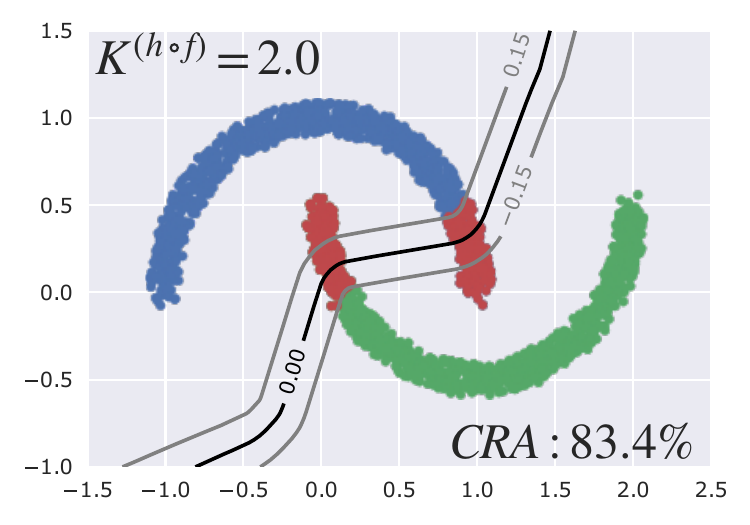}
    \end{subfigure}
    \hfill
    
    \begin{minipage}[t]{0.014\textwidth}
        \vspace{2em}
        \rotatebox[origin=c]{90}{$\gamma = 1.0$}
    \end{minipage}
    \begin{subfigure}[t]{0.24\textwidth}
        \centering
        \vspace{-0.1em}
        \includegraphics[width=\textwidth]{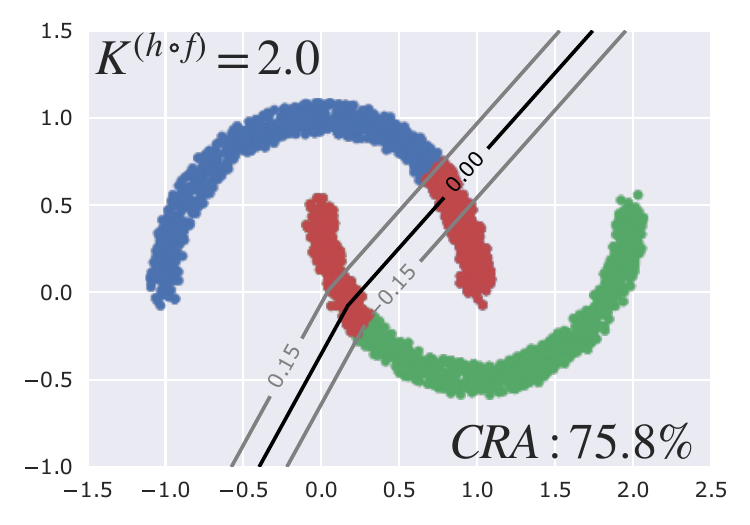}
    \end{subfigure}
    \begin{subfigure}[t]{0.24\textwidth}
        \centering
        \vspace{-0.1em}
        \includegraphics[width=\textwidth]{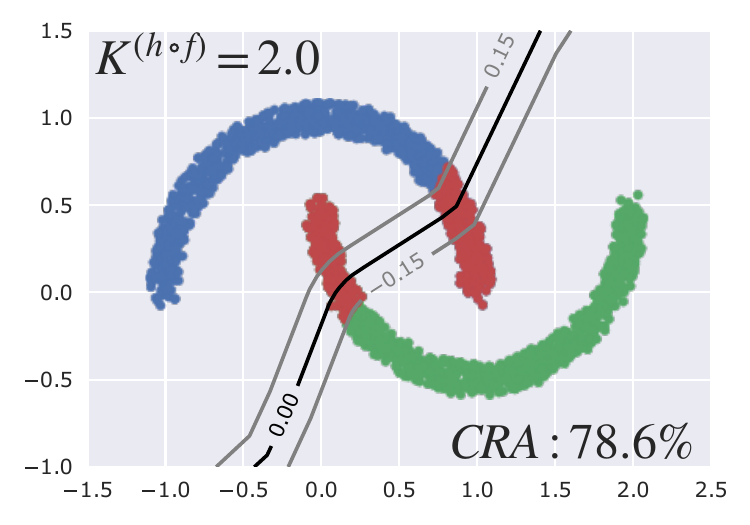}
    \end{subfigure}
    \begin{subfigure}[t]{0.24\textwidth}
        \centering
        \vspace{-0.1em}
        \includegraphics[width=\textwidth]{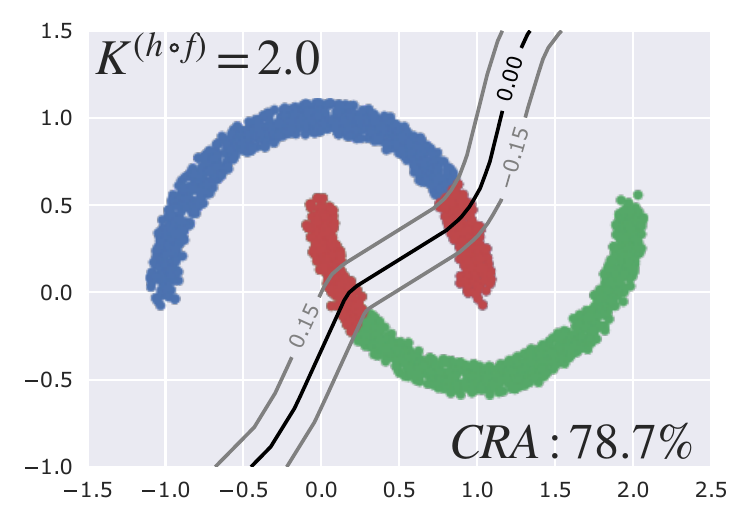}
    \end{subfigure}
    \hfill
    \caption{1-Lipschitz constrained SOC is too smooth to fit two-moons, irrespective of depth or its margin controlling parameter $\gamma$. All networks use last-layer normalization~\cite{soc}, which increases $\khf$ to $2.0$}
    \label{fig:appendix:moons_soc}
\end{figure*}

\myparagraph{Faster convergence with \ac{ours}.}
For training deep models we highlight an additional benefit provided by \ac{ours} that is of practical importance: \textbf{faster convergence}.
We plot the certified robust accuracy (CRA) over number of epochs in figure~\ref{fig:convergence} and find that our loss converges faster and is less noisy over the alternative large margin loss: GloRo.
Note that in this instance, we train GloRo with a fixed $\epsilon$ schedule to draw a fair comparison.

\begin{figure*}
    \centering
    \begin{subfigure}[t]{0.32\textwidth}
        \includegraphics[width=\textwidth]{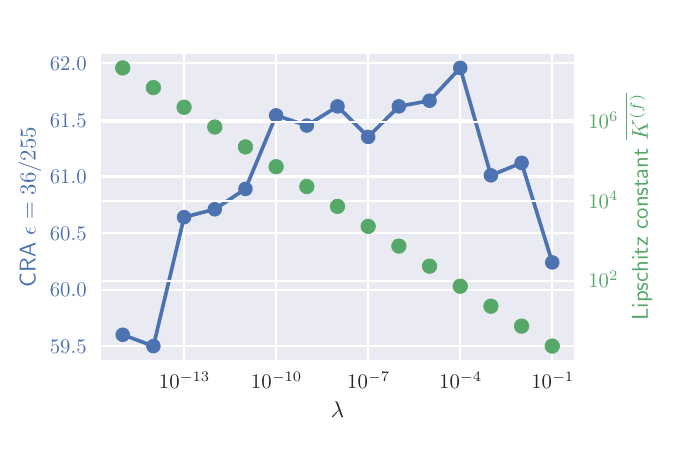}
        \caption{\ac{ours} is insensitive to a broad range of $\lambda$ choices.}
        \label{fig:lambda_sweep}
    \end{subfigure}
    \hfill
    \begin{subfigure}[t]{0.32\textwidth}
         \includegraphics[width=\textwidth, trim={0.3cm 0 0.2cm 0.2cm},clip]{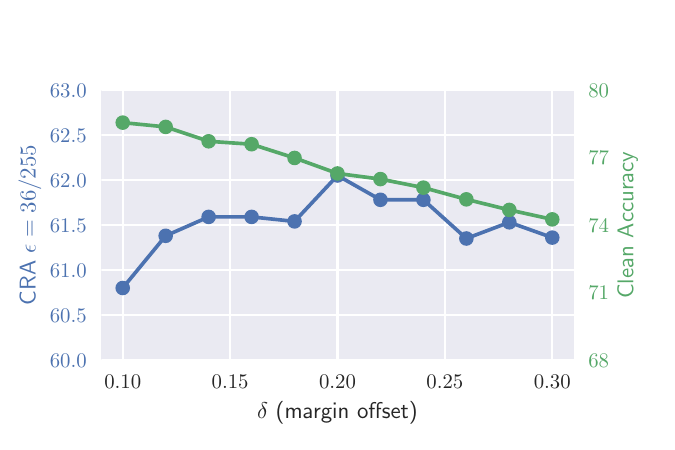}
         \caption{Tuning $\delta$ can improve CRA further.}
         \label{fig:appendix:offset_sweep}
    \end{subfigure}
    \hfill
    \begin{subfigure}[t]{0.32\textwidth}
        \includegraphics[width=\textwidth]{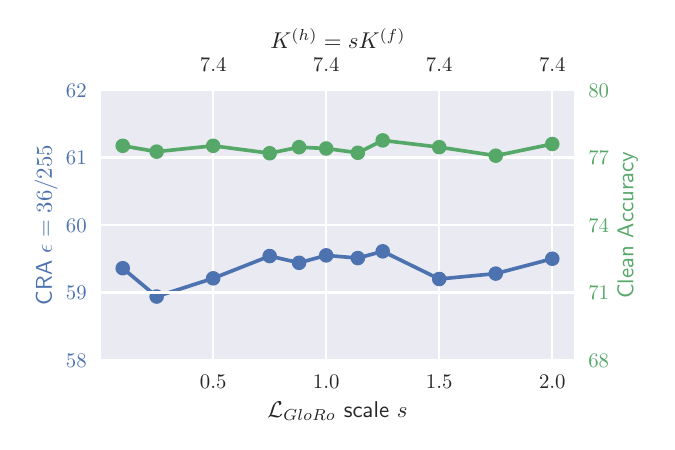}
        \caption{Na\"ively scaling outputs in GloRo has no effect on $\khf$ (top x-axis) or CRA (blue).}
        \label{fig:gloro_scale_sweep}
    \end{subfigure}
    \caption{Hyperparameter sweeps for $\lambda$ (left) and margin offset $\delta$ (middle) evaluated on CIFAR-10. $\lambda$ specifies minimization pressure on $\kf$ and $\delta$ can be tuned beyond the canonical value to increase robust accuracy. The right figure shows that GloRo is not endowed with \ac{ours}s slack property.}
    \label{fig:param_sweep}
\end{figure*}

\myparagraph{Regularizing coefficient $\lambda$.}
As discussed in section~3.3 (main paper), $\lambda$ is required to add slight regularization pressure on $\kf$ -- the Lipschitz constant of the classifier $f$.
We subsequently investigate the impact on CRA when choosing different values for $\lambda$.
For this experiment, we use the \textit{4C3F} architecture and train on CIFAR-10.
Figure~\ref{fig:lambda_sweep} shows CRA in blue (left vertical axis) and Lipschitz constant $\overline{K}$ in green (right axis) over parameter $\lambda$ which is increased from $10^{-15}$ to $10^{-1}$.
Note that $\kf$ rises exponentially with decreasing $\lambda$.
We observe a wide range of values leading to state-of-the-art CRA results ($10^{-10} \leq \lambda \leq 10^{-4}$), indicating that CRA is fairly insensitive to the choice of $\lambda$.
Yet we also see two substantial drops in performance when $\lambda$ is set too high ($\geq 10^{-3}$) or too low $\leq 10^{-11}$).
Choosing $\lambda$ too high leads to reduced complexity in $f$, which is undesired.
On the other hand, it is not directly evident why choosing $\lambda$ too low also impairs performance.
This observation necessitates further investigation in future work.

\myparagraph{$\epsilon$ -- $p$ ablation.}
We showed in the main paper, that slack $p$ can be used to trade-off clean and robust accuracies on \textit{4C3F} on CIFAR-10 for a fixed $\epsilon_\text{train}$.
For completeness, we extend this experiment with the question: can we maximize the margin while retaining clean accuracy by reducing slack.
We evaluate combinations of $\epsilon_\text{train}$ and slack on \textit{8C2F} on Tiny-ImageNet.
Results for clean accuracies and CRA for $\epsilon=36/255$ and $\epsilon=108/255$ is presented in table~\ref{tab:appendix:eps_p_matrix}.
Generally, we observe a similar trend to the main paper results.
With decreasing $p$ and $\epsilon$, we observe an increase of clean accuracy while CRA decreases and in reverse: a decrease in clean accuracy while CRA increases with increasing $p$ and $\epsilon$.
Importantly though, we can indeed find a configuration that increases the margin, improves CRA and retains clean accuracy.
Take $\epsilon=0.5$ and $p=0.05$ for example.
This configuration achieves $37.1\%$ clean and $25.7\%$ and $12.0\%$ robust accuracies.
We can find a configuration with higher $\epsilon$ but most importantly lower $p$ that outperforms all metrics: $\epsilon=0.75$ and $p=0.005$.
This configuration achieves $37.5\% (+0.4)$ clean accuracy and $26.0\% (+0.3)$ and $12.0\%$ robust accuracies.
And, as discussed in section~3.3 (main paper), we also find the better configuration to obtain a higher Lipschitz constant, \ie{} $\khf=8.0$ vs $\khf=7.4$.
Consequently, training for larger $\epsilon$ margins can still result in overall better clean and robust accuracies when the slack is reduced appropriately.
It needs to be seen, whether this relationship can be harnessed further for additional improvements.

\begin{table}[htb]
\centering
\caption{Clean and certified robust accuracies for $25$ combinations of $\epsilon_\text{train}$ and $p$ for \ac{ours} on \textit{8C2F} on Tiny-ImageNet. The numbers shown are all trained with a random seed value of $1$.}
\label{tab:appendix:eps_p_matrix}
\begin{tabular}{|l|c|c|c|c|c||c|c|c|c|c||c|c|c|c|c|}
\hline
 & \multicolumn{5}{c||}{Clean accuracy} & \multicolumn{5}{c||}{CRA 36/255} & \multicolumn{5}{c|}{CRA 108/255} \\
$p$ \textbackslash\, $\epsilon_\text{train}$ & 0.15 & 0.25 & 0.5 & 0.75 & 1.0 & 0.15 & 0.25 & 0.5 & 0.75 & 1.0 & 0.15 & 0.25 & 0.5 & 0.75 & 1.0 \\ \hline
0.1 & 43.7 & 40.3 & 36.2 & 32.4 & 30.4 & 22.5 & 25.2 & 25.9 & 25.1 & 24.1 & 4.9 & 8.7 & 12.8 & 14.1 & 14.8 \\ \hline
0.05 & 44.2 & 42.0 & 37.1 & 34.5 & 32.1 & 21.8 & 25.1 & 25.7 & 26.0 & 25.2 & 4.2 & 7.9 & 12.0 & 13.7 & 14.5 \\ \hline
0.01 & 45.4 & 43.0 & 39.7 & 36.4 & 35.1 & 20.0 & 23.4 & 25.9 & 25.6 & 25.7 & 3.4 & 6.7 & 10.7 & 12.4 & 13.2 \\ \hline
0.005 & 44.4 & 43.4 & 39.5 & 37.5 & 35.5 & 19.5 & 23.2 & 25.5 & 26.0 & 26.0 & 2.8 & 6.5 & 10.4 & 12.0 & 12.9 \\ \hline
0.001 & 45.3 & 43.9 & 40.7 & 38.5 & 37.2 & 18.5 & 21.7 & 24.8 & 25.5 & 25.9 & 2.7 & 5.8 & 9.4 & 11.1 & 12.0 \\\hline
\end{tabular}
\begin{tabular}{|l|c|c|c|c|c|}
\hline
 & \multicolumn{5}{c|}{$\khf$} \\
$p$ \textbackslash\, $\epsilon_\text{train}$ & 0.15 & 0.25 & 0.5 & 0.75 & 1.0 \\ \hline
0.1 & 19.6 & 11.8 & 5.9 & 3.9 & 2.9 \\ \hline
0.05 & 24.4 & 14.7 & 7.4 & 4.9 & 3.7 \\ \hline
0.01 & 35.3 & 21.2 & 10.6 & 7.1 & 5.3 \\ \hline
0.005 & 40.0 & 24.0 & 12.0 & 8.0 & 6.0 \\ \hline
0.001 & 50.7 & 30.4 & 15.2 & 10.1 & 7.6 \\\hline
\end{tabular}
\end{table}

\myparagraph{Comparison to fixed temperature scaling.}
We note that scaling the outputs of existing margin losses is not sufficient to control slack -- and thus $\khf$, as the optimization is able to compensate for this factor.
We run a simple experiment utilizing GloRo and varying temperature $s=\nicefrac{1}{T}$ in their loss $\mathcal{L}_\text{GloRo}(y; h(s \cdot f(x)))$ and present results in figure~\ref{fig:gloro_scale_sweep}.
In contrast to our loss, we observe that such scaling (bottom x-axis) has no impact on $\khf$ (top x-axis) and little to no impact on accuracies.

\myparagraph{Margin offset as tuneable parameter.}
In our \ac{ours} definition in the main paper, we add or subtract $\epsilon$ as hard margin offsets to improve margin training.
We can consider this offset as hyper-parameter $\delta$, such that the calibrated logistic is now defined as:
\begin{align*}
    \hat{h}(f(x);y) = h\left(-\frac{y\delta}{\sigma_\epsilon(p)} + \frac{1}{\sigma_{\epsilon}(p)}\frac{f(x)}{K}\right).
\end{align*}
Instead of the canonical value $\delta=\epsilon$, we choose $11$ values between $0.1$ and $0.3$ and plot robust and clean accuracy in figure~\ref{fig:appendix:offset_sweep}.
While robust accuracy (blue curve) can be improved with greater $\delta$, clean accuracy consistently decreases.
Here, $\delta=0.2$ achieves a robust accuracy of $62.0\%$, improving on our main results ($+0.6\%$).
Concurrently, clean accuracy decreases to $76.2\% (-1.4\%)$.

\subsection{Main results under different random seeds}
To the main results in table~1 and 2 (main paper), we here report all runs under different seeds when using~\ac{ours} in table~\ref{tab:appendix:cifar10_seeds} (CIFAR-10) and table~\ref{tab:appendix:tiny_imagenet_seeds} (Tiny-ImageNet).
In the last column, we report additionally, the average value and its standard deviation.
We highlight, that the standard deviation is similar to reported values in related work.

\begin{table*}[ht]
\centering
\caption{\ac{ours} results on CIFAR-10 starting from different random seeds. }
\label{tab:appendix:cifar10_seeds}
\resizebox{\linewidth}{!}{
\begin{tabular}{|l|l|c|c|c|c|c|c|c|c|c||c|}
\hline
\multirow{2}{*}{Model} & \multirow{2}{*}{Metric} & \multicolumn{9}{c||}{Random seed value} & \multirow{2}{*}{Avg} \\\cline{3-11}
 &  & 1 & 2 & 3 & 4 & 5 & 6 & 7 & 8 & 9 &  \\\hline\hline
\multirow{6}{*}{\ac{ours}@4C3F} 
 & Acc & 77.3 & 77.1 & 77.5 & 77.3 & 77.3 & 77.5 & 77.3 & 77.6 & 77.2 & 77.3 $\pm$ 0.2 \\\cline{2-12}
 & CRA 36/255 & 61.2 & 61.5 & 61.3 & 61.2 & 61.6 & 61.7 & 61.5 & 61.5 & 61.4 & 61.0 $\pm$ 0.2 \\\cline{2-12}
 & CRA 72/255 & 44.5 & 44.5 & 44.1 & 43.6 & 44.1 & 44.5 & 44.5 & 44.0 & 44.2 & 44.2 $\pm$ 0.3 \\\cline{2-12}
 & CRA 108/255 & 29.3 & 29.1 & 29.0 & 28.7 & 29.1 & 29.2 & 29.3 & 29.1 & 29.0 & 29.1 $\pm$ 0.2 \\\cline{2-12}
 & $\kf$ & 72.1 & 72.0 & 72.0 & 72.0 & 72.2 & 72.2 & 72.2 & 71.8 & 72.1 & 72.1 $\pm$ 0.2\\\cline{2-12}
 & Tightness & 80.6 & 79.5 & 81.2 & 80.8 & 79.5 & 79.8 & 79.8 & 80.8 & 79.6 & 80.2 $\pm$ 0.7\\\hline\hline
\multirow{6}{*}{\ac{ours}@6C2F} 
 & Acc & 77.5 & 77.8 & 77.6 & 77.7 & 77.4 & 77.8 & 77.5 & 77.6 & 77.9 & 77.6 $\pm$ 0.2 \\\cline{2-12}
 & CRA 36/255 & 61.3 & 61.1 & 61.2 & 61.4 & 61.1 & 61.1 & 61.2 & 61.7 & 61.4 & 61.3 $\pm$ 0.2 \\\cline{2-12}
 & CRA 72/255 & 43.4 & 43.7 & 43.6 & 43.5 & 43.5 & 43.3 & 43.4 & 43.7 & 43.6 & 43.5 $\pm$ 0.1 \\\cline{2-12}
 & CRA 108/255 & 27.6 & 27.8 & 27.8 & 27.7 & 27.2 & 27.6 & 27.6 & 28.1 & 28.1 & 27.7 $\pm$ 0.3 \\\cline{2-12}
 & $\kf$ & 709.3 & 704.3 & 704.6 & 708.8 & 722.5 & 710.2 & 708.1 & 707.1 & 709.1 & 709.3 $\pm$ 5.3 \\\cline{2-12}
 & Tightness & 76.6 & 78.2 & 77.4 & 78.7 & 76.3 & 77.4 & 77.5 & 77.9 & 78.3 & 77.6 $\pm$ 0.8\\\hline\hline
\multirow{6}{*}{\ac{ours}@LipConv-20}
 & Acc & 77.4 & 76.8 & 77.8 & 77.4 & 77.3 & 77.9 & 77.3 & 77.1 & 77.7 & 77.4 $\pm$ 0.3\\ \cline{2-12}
 & CRA 36/255 & 64.1 & 64.0 & 64.2 & 64.1 & 64.4 & 64.4 & 64.4 & 63.8 & 64.4 & 64.2 $\pm$ 0.2 \\ \cline{2-12}
 & CRA 72/255 & 49.5 & 49.4 & 49.4 & 49.7 & 49.4 & 49.8 & 49.7 & 49.0 & 49.9 & 49.5 $\pm$ 0.2 \\ \cline{2-12}
 & CRA 108/255 & 37.3 & 36.3 & 36.4 & 36.5 & 36.5 & 36.8 & 37.1 & 36.5 & 36.7 & 36.7 $\pm$ 0.3 \\ \cline{2-12}
 & $\kf$ & 35.6 & 35.7 & 35.7 & 35.7 & 35.5 & 36.0 & 35.8 & 35.9 & 35.9 & 35.8 $\pm$ 0.2 \\ \cline{2-12}
 & Tightness & 85.8 & 85.6 & 85.3 & 86.3 & 85.1 & 85.7 & 85 & 85.6 & 86.3 & 85.6 $\pm$ 0.5 \\\hline\hline
\multirow{6}{*}{\ac{ours}@CPL-XL} 
 & Acc & 78.9 & 79.0 & 78.8 & 78.5 & 78.9 & 78.8 & 78.6 & 78.7 & 78.7 & 78.8 $\pm$ 0.1 \\\cline{2-12}
 & CRA 36/255 & 66.0 & 66.0 & 66.0 & 65.7 & 65.8 & 66.0 & 65.8 & 65.7 & 65.9 & 65.9 $\pm$ 0.1 \\\cline{2-12}
 & CRA 72/255 & 51.6 & 51.6 & 51.5 & 51.7 & 51.6 & 51.9 & 51.5 & 51.7 & 51.7 & 51.6 $\pm$ 0.1 \\\cline{2-12}
 & CRA 108/255 & 38.1 & 38.2 & 37.9 & 37.8 & 38.0 & 38.0 & 38.3 & 38.2 & 38.2 & 38.1 $\pm$ 0.2 \\\cline{2-12}
 & $\kf$ & 34.6 & 34.6 & 34.6 & 34.6 & 34.6 & 34.5 & 34.6 & 34.6 & 34.5 & 34.5 $\pm$ 0.1 \\\cline{2-12}
 & Tightness & 80.5 & 80.6 & 81.7 & 80.1 & 80.3 & 80.1 & 79.8 & 80.2 & 79.9 & 80.4 $\pm$ 0.5 \\\hline\hline
\multirow{6}{*}{\ac{ours}@SLL-XL} 
 & Acc & 73.0 & 72.9 & 73.1 & 73.2 & 72.9 & 73.1 & 73.0 & 72.8 & 73.1 & 73.0 $\pm$ 0.1\\\cline{2-12}
 & CRA 36/255 & 65.5 & 65.5 & 65.3 & 65.7 & 65.4 & 65.5 & 65.4 & 65.5 & 65.6 & 65.5 $\pm$ 0.1 \\\cline{2-12}
 & CRA 72/255 & 57.8 & 57.6 & 57.9 & 57.7 & 57.5 & 57.7 & 57.7 & 57.9 & 58.2 & 57.8 $\pm$ 0.2 \\\cline{2-12}
 & CRA 108/255 & 51.0 & 50.9 & 50.7 & 51.2 & 50.9 & 50.8 & 51.0 & 51.1 & 51.0 & 51.0 $\pm$ 0.1\\\cline{2-12}
 & $\kf$ & 58.9 & 57.3 & 58.3 & 60.0 & 61.1 & 62.1 & 60.0 & 57.7 & 59.1 & 59.4 $\pm$ 1.6 \\\cline{2-12}
 & Tightness & 87.0 & 89.0 & 88.0 & 87.8 & 87.9 & 88.2 & 87.7 & 88.3 & 88.0 & 88.0 $\pm$ 0.5 \\\hline\hline
\end{tabular}
}
\end{table*}

\begin{table*}[ht]
\centering
\caption{\ac{ours} results on CIFAR-100 starting from different random seeds. }
\label{tab:appendix:cifar100_seeds}
\resizebox{\linewidth}{!}{
\begin{tabular}{|l|l|c|c|c|c|c|c|c|c|c||c|}
\hline
\multirow{2}{*}{Model} & \multirow{2}{*}{Metric} & \multicolumn{9}{c||}{Random seed value} & \multirow{2}{*}{Avg} \\\cline{3-11}
 &  & 1 & 2 & 3 & 4 & 5 & 6 & 7 & 8 & 9 &  \\\hline\hline
\multirow{6}{*}{\ac{ours}@LipConv-20} 
 & Acc & 48.8 & 48.4 & 48.4 & 48.0 & 48.0 & 48.5 & 48.4 & 47.3 & 47.9 & 48.2 $\pm$ 0.4 \\\cline{2-12}
 & CRA 36/255 & 35.4 & 35.2 & 35.2 & 35.4 & 35.0 & 35.2 & 35.2 & 34.3 & 35.1 & 35.1 $\pm$ 0.3 \\\cline{2-12}
 & CRA 72/255 & 25.6 & 25.4 & 25.5 & 25.3 & 25.5 & 25.6 & 25.3 & 24.4 & 25.2 & 25.3 $\pm$ 0.3 \\\cline{2-12}
 & CRA 108/255 & 18.6 & 18.4 & 18.4 & 18.3 & 18.4 & 18.7 & 18.1 & 17.7 & 18.4 & 18.3 $\pm$ 0.3 \\\cline{2-12}
 & $\kf$ & 45.3 & 45.5 & 45.6 & 45.3 & 45.6 & 45.5 & 45.2 & 45.3 & 45.7 & 45.4 $\pm$ 0.2 \\\cline{2-12}
 & Tightness & 84.4 & 84.1 & 85.2 & 83.2 & 83.5 & 85.3 & 83.8 & 85.0 & 84.6 & 84.3 $\pm$ 0.7 \\\hline\hline
\multirow{6}{*}{\ac{ours}@CPL-XL} 
 & Acc & 48.0 & 47.8 & 48.0 & 48.2 & 47.9 & 47.8 & 47.7 & 48.2 & 47.9 & 47.9 $\pm$ 0.2 \\\cline{2-12}
 & CRA 36/255 & 36.1 & 36.7 & 36.4 & 36.1 & 36.5 & 36.1 & 36.5 & 36.0 & 36.3 & 36.3 $\pm$ 0.2 \\\cline{2-12}
 & CRA 72/255 & 28.0 & 27.9 & 28.0 & 28.9 & 28.0 & 28.0 & 28.0 & 27.7 & 28.0 & 28.1 $\pm$ 0.3 \\\cline{2-12}
 & CRA 108/255 & 21.5 & 21.5 & 21.8 & 21.5 & 21.5 & 21.5 & 21.4 & 21.4 & 21.7 & 21.5 $\pm$ 0.1 \\\cline{2-12}
 & $\kf$ & 42.0 & 42.0 & 42.0 & 41.9 & 42.0 & 42.0 & 42.0 & 42.0 & 42.2 & 42.0 $\pm$ 0.1 \\\cline{2-12}
 & Tightness & 79.0 & 79.4 & 78.1 & 78.8 & 79.0 & 77.7 & 79.9 & 78.6 & 77.9 & 78.7 $\pm$ 0.7 \\\hline\hline
\multirow{6}{*}{\ac{ours}@SLL-XL} 
 & Acc & 47.1 & 46.9 & 46.9 & 46.8 & 47 & 47.1 & 46.6 & 46.6 & 47.4 & 46.9 $\pm$ 0.3 \\\cline{2-12}
 & CRA 36/255 & 36.9 & 36.7 & 36.3 & 36.6 & 36.5 & 36.6 & 36.5 & 36.7 & 36.7 & 36.6 $\pm$ 0.2 \\\cline{2-12}
 & CRA 72/255 & 29 & 29.2 & 29 & 28.9 & 29 & 29.2 & 29 & 29 & 29 & 29.0 $\pm$ 0.1 \\\cline{2-12}
 & CRA 108/255 & 23.6 & 23.6 & 23.5 & 23.4 & 23.8 & 23.3 & 23.4 & 23.3 & 23.1 & 23.4 $\pm$ 0.2 \\\cline{2-12}
 & $\kf$ & 1.2 & 1.5 & 1.2 & 1.2 & 1.2 & 1.3 & 1.2 & 1.2 & 1.2 & 1.3 $\pm$ 0.1\\\cline{2-12}
 & Tightness & 79.8 & 79.5 & 79.5 & 80.4 & 79.1 & 82.0 & 78.5 & 81.5 & 80.5 & 80.1 $\pm$ 1.1\\\hline
\end{tabular}
}
\end{table*}

\begin{table*}[ht]
\centering
\caption{\ac{ours} results on Tiny-ImageNet starting from different random seeds.}
\label{tab:appendix:tiny_imagenet_seeds}
\resizebox{\linewidth}{!}{
\begin{tabular}{|l|l|c|c|c|c|c|c|c|c|c||c|}
\hline
\multirow{2}{*}{Model} & \multirow{2}{*}{Metric} & \multicolumn{9}{c||}{Random seed value} & \multirow{2}{*}{Avg} \\\cline{3-11}
 &  & 1 & 2 & 3 & 4 & 5 & 6 & 7 & 8 & 9 &  \\\hline\hline
\multirow{6}{*}{$\begin{matrix}\text{\ac{ours}@8C2F}\\ \epsilon=0.5, p=0.01\end{matrix}$} 
 & Acc & 39.7 & 39.8 & 39.6 & 39.6 & 40.0 & 39.8 & 39.6 & 39.9 & 39.9 & 39.8 $\pm$ 0.1 \\ \cline{2-12}
 & CRA 36/255 & 25.5 & 25.7 & 25.7 & 26.3 & 26.1 & 25.9 & 25.5 & 26.1 & 26.3 & 25.9 $\pm$ 0.3 \\ \cline{2-12}
 & CRA 72/255 & 16.3 & 16.5 & 16.5 & 16.7 & 16.4 & 16.7 & 16.5 & 16.7 & 16.6 & 16.5 $\pm$ 0.1\\ \cline{2-12}
 & CRA 108/255 & 10.7 & 10.7 & 10.6 & 10.6 & 10.9 & 10.9 & 10.8 & 10.7 & 10.7 & 10.7 $\pm$ 0.1 \\ \cline{2-12}
 & $\kf$ & 287.7 & 288.6 & 287.9 & 287.7 & 291.0 & 291.5 & 285.5 & 288.4 & 286.7 & 288.3 $\pm$ 1.9 \\ \cline{2-12}
 & Tightness & 63.6 & 64.8 & 64.4 & 64.1 & 63.9 & 64.5 & 64.0 & 64.9 & 64.7 & 64.2 $\pm$ 0.5 \\ \hline\hline
\multirow{6}{*}{$\begin{matrix}\text{\ac{ours}@8C2F}\\ \epsilon=1.0, p=0.025\end{matrix}$} 
 & Acc & 33.6 & 33.3 & 33.5 & 33.0 & 33.4 & 33.1 & 33.6 & 33.9 & 33.7 & 33.5 $\pm$ 0.3\\ \cline{2-12}
 & CRA 36/255 & 25.6 & 24.9 & 25.4 & 25.4 & 25.2 & 25.3 & 25.1 & 25.3 & 25.3 & 25.3 $\pm$ 0.2\\ \cline{2-12}
 & CRA 72/255 & 19.2 & 18.6 & 19.0 & 19.0 & 18.9 & 18.9 & 18.9 & 19.1 & 19.0 & 19.0 $\pm$ 0.2\\ \cline{2-12}
 & CRA 108/255 & 13.9 & 13.3 & 13.9 & 13.6 & 13.9 & 14.1 & 14.1 & 13.9 & 13.7 & 13.8 $\pm$ 0.2\\ \cline{2-12}
 & $\kf$ & 24.5 & 24.3 & 24.5 & 24.4 & 24.4 & 24.4 & 24.5 & 24.4 & 24.3 & 24.4 $\pm$ 0.1\\ \cline{2-12}
 & Tightness & 73.5 & 71.2 & 72.9 & 72.1 & 73.8 & 71.2 & 74.1 & 73.7 & 72.7 & 72.8 $\pm$ 1.0\\ \hline\hline
\multirow{6}{*}{\ac{ours}@LipConv-10} & Acc & 32.5 & 31.8 & 32.8 & 32.1 & 31.6 & 33.4 & 32.4 & 32.6 & 32.2 & 32.4 $\pm$ 0.5 \\\cline{2-12}
 & CRA 36/255 & 25.4 & 25.2 & 24.8 & 25.0 & 24.4 & 24.9 & 25.1 & 24.9 & 25.0 & 25.0 $\pm$ 0.3 \\\cline{2-12}
 & CRA 72/255 & 18.9 & 18.8 & 18.4 & 18.4 & 17.2 & 18.3 & 18.8 & 18.5 & 18.4 & 18.4 $\pm$ 0.5 \\\cline{2-12}
 & CRA 108/255 & 14.0 & 14.1 & 13.1 & 13.1 & 11.6 & 13.4 & 13.8 & 13.5 & 13.4 & 13.3 $\pm$ 0.7 \\\cline{2-12}
 & $\kf$ & 6.8 & 6.1 & 6.3 & 7.3 & 6.8 & 6.3 & 6.9 & 6.8 & 6.8 & 6.7 $\pm$ 0.4 \\\cline{2-12}
 & Tightness & 84.1 & 81.3 & 81.8 & 84.0 & 82.3 & 82.0 & 81.2 & 83.8 & 81.8 & 82.5 $\pm$ 1.2 \\\hline
\multirow{6}{*}{\ac{ours}@LipConv-20} & Acc & 33.4 & 32.7 & 32.5 & 32.5 & 32.1 & 31.7 & 32.8 & 33.0 & 32.6 & 32.6 $\pm$ 0.5 \\\cline{2-12}
 & CRA 36/255 & 25.9 & 26.2 & 26.1 & 26.2 & 25.6 & 26.1 & 26.4 & 26.0 & 25.8 & 26.0 $\pm$ 0.2 \\\cline{2-12}
 & CRA 72/255 & 19.7 & 20.4 & 20.5 & 20.4 & 20.1 & 20.3 & 20.3 & 20.3 & 20.1 & 20.2 $\pm$ 0.2 \\\cline{2-12}
 & CRA 108/255 & 14.5 & 16.0 & 15.6 & 15.6 & 15.5 & 15.8 & 15.9 & 15.3 & 15.5 & 15.5 $\pm$ 0.4 \\\cline{2-12}
 & $\kf$ & 4.0 & 5.2 & 5.0 & 5.0 & 5.0 & 5.4 & 4.9 & 4.6 & 4.9 & 4.9 $\pm$ 0.4 \\\cline{2-12}
 & Tightness & 81.9 & 84.5 & 84.4 & 84.2 & 82.9 & 84.5 & 83.9 & 87.1 & 85.1 & 84.3 $\pm$ 1.4 \\\hline
\end{tabular}
}
\end{table*}

\bibliographystyle{splncs04}
\bibliography{supplement}